\documentclass[sigconf]{acmart}
\usepackage{xcolor}
\usepackage{subcaption}
\usepackage{multirow}

\usepackage{algorithm}          
\usepackage{algorithmic}
\usepackage{bm}

\usepackage{graphicx}

\def\V#1{\boldsymbol{\mathit#1}}          

\AtBeginDocument{%
  }

\copyrightyear{2025}
\acmYear{2025}
\setcopyright{acmlicensed}\acmConference[MM '25]{Proceedings of the 33rd ACM International Conference on Multimedia}{October 27--31, 2025}{Dublin, Ireland}
\acmBooktitle{Proceedings of the 33rd ACM International Conference on Multimedia (MM '25), October 27--31, 2025, Dublin, Ireland}
\acmDOI{10.1145/3746027.3754719}
\acmISBN{979-8-4007-2035-2/2025/10}

\begin{document}

\title{DSS-Prompt: Dynamic-Static Synergistic Prompting for Few-Shot Class-Incremental Learning}

\author{Linpu He}
\affiliation{%
  \institution{Department of Computer Science and Technology, Zhejiang University}
  \city{Hangzhou}
  \state{Zhejiang}
  \country{China}
}
\email{help@zju.edu.cn}

\author{Yanan Li}
\authornote{Corresponding Author}
\affiliation{%
  \institution{Research Center for Frontier Fundamental Studies, Zhejiang Lab}
  \city{Hangzhou}
  \state{Zhejiang}
  \country{China}
}
\email{liyn@zhejianglab.com}

\author{Bingze Li}
\affiliation{%
  \institution{Department of Computer Science and Technology, Zhejiang University}
  \city{Hangzhou}
  \state{Zhejiang}
  \country{China}
}
\email{libz@zju.edu.cn}

\author{Elvis Han Cui}
\affiliation{%
  \institution{Department of Neurology, University of California, Irvine}
  \city{Irvine}
  \state{CA}
  \country{USA}
}
\email{elviscuihan@g.ucla.edu}

\author{Donghui Wang}
\affiliation{%
  \institution{Department of Computer Science and Technology, Zhejiang University}
  \city{Hangzhou}
  \state{Zhejiang}
  \country{China}
}
\email{dhwang@zju.edu.cn}

\renewcommand{\shortauthors}{Linpu He, Yanan Li, Bingze Li, Elvis Han Cui, \& Donghui Wang}
\begin{abstract}
Learning from large-scale pre-trained models with strong generalization ability has shown remarkable success in a wide range of downstream tasks recently, but it is still underexplored in the challenging few-shot class-incremental learning (FSCIL) task. It aims to continually learn new concepts from limited training samples without forgetting the old ones at the same time. In this paper, we introduce DSS-Prompt, a simple yet effective approach that transforms the pre-trained Vision Transformer with minimal modifications in the way of prompts into a strong FSCIL classifier. Concretely, we synergistically utilize two complementary types of prompts in each Transformer block: static prompts to bridge the domain gap between the pre-training and downstream datasets, thus enabling better adaption; and dynamic prompts to capture instance-aware semantics, thus enabling easy transfer from base to novel classes. Specially, to generate dynamic prompts, we leverage a pre-trained multi-modal model to extract input-related diverse semantics, thereby generating complementary input-aware prompts, and then adaptively adjust their importance across different layers. 
In this way, on top of the prompted visual embeddings, a simple prototype classifier can beat state-of-the-arts without further training on the incremental tasks. We conduct extensive experiments on four benchmarks to validate the effectiveness of our DSS-Prompt and show that it consistently achieves better performance than existing approaches on all datasets and can alleviate the catastrophic forgetting issue as well. 
\end{abstract}

\ccsdesc[500]{Computing methodologies~Image representations}

\ccsdesc[500]{Computing methodologies~Computer vision representations}

\keywords{few-shot class-incremental learning, prompt tuning, dynamic prompts, instance-aware multi-modal knowledge}


\maketitle

\section{Introduction}
\label{sec:introduction}

\begin{figure}[htbp]
    \centering
    \includegraphics[width=0.45\textwidth]{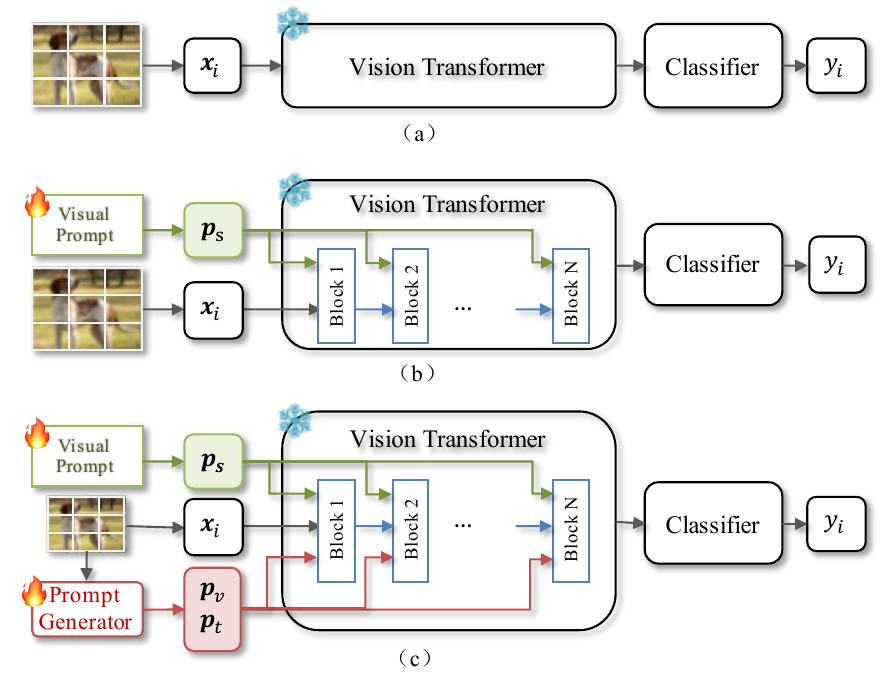}
    \caption{Comparison between (a) the naive FSCIL baseline based on the pre-trained ViT, (b) existing FSCIL paradigm that adds prompts into several or every block in ViT, and (c) our DSS-Prompt that synergistically uses both dynamic and static prompts to capture both domain-specific and instance-aware features, enabling the well balance between generalization and adaptation in FSCIL. }
    \Description{fig:Baselines}
    \label{fig:Baselines}
\end{figure}

Deep neural networks have made remarkable progress across various applications, largely due to the availability of vast training data, advanced architectures, and computational resources at their disposal. These achievements typically rely on having access to all the required data at once for training on different tasks. However, in real-world applications, e.g. face recognition~\cite{peng2022few} and medical diagnosis~\cite{derakhshani2022lifelonger}, new object categories continuously emerge and collecting large annotated datasets for every new class is impractical. This requires deep neural networks to continually learn new classes with only a few labeled examples while retaining knowledge of the previously learned ones, i.e., few-shot class-incremental learning (FSCIL)~\cite{tao2020few, agarwal2022semantics, park2024pre, liu2025few, zhang2025few}.

FSCIL typically begins with a base session, where the model is trained on abundant samples from base classes, followed by multiple incremental sessions that provide only a few samples per new class, requiring the model to quickly generalize to these classes. Throughout the continual learning process, two notorious challenges occur: overfitting and catastrophic forgetting, which erases the previously learned knowledge and is further exacerbated by the former. 
To address the above issues, a mainstream strategy in FSCIL is to learn feature embeddings during the base session that not only generalize well, but also retrain strong discrimination ability. By doing so, it enables the model to achieve a good balance between model stability and plasticity~\cite{zhang2025few}, even without fine-tuning on new data~\cite{zhang2021few, wang2024few, ahmed2024orco}. For example, ~\cite{zhang2021few} decouples the learning of feature embeddings and classifiers and proposes to optimize the representations only in the base session, while ~\cite{ahmed2024orco} assigns orthogonal features to incoming new classes in advance. These prior efforts are mainly based on the premise of learning from scratch, attempting to mitigate the forgetting issue when adapting to each incremental task, and have achieved significant progresses. 

Recently, pre-trained Vision Transformers (ViTs) have shown impressive zero-shot abilities across a variety of downstream tasks including image classification~\cite{han2022survey}, detection~\cite{jia2024purified}, and segmentation~\cite{ghiasi2022scaling, yu2024pffaa}. These models are trained using web-scale data and offer unique representations with strong generalization ability and transfer capabilities, therefore inspiring the FSCIL community to design solutions based on it. Naively applying the pre-trained ViT in FSCIL, as shown in Fig.\ref{fig:Baselines} (a), is ineffective, due to the explicit distribution gap between upstream and downstream datasets~\cite{park2024pre}. A more efficient strategy is fixing the backbone parameters in ViTs and only training a few parameters through prompts, as shown in Fig.\ref{fig:Baselines} (b), along with incorporating some typical strategies (e.g., replay) to prevent the forgetting issue~\cite{huang2024learning, park2024pre, li2024few, liu2025few}. For example, ~\cite{park2024pre} selectively trains specific layers with a new prompt modulation approach and adds language-guided knowledge distillations, while ~\cite{liu2025few} splits prompts into bridged task-invariant and task-specific prompts to enable knowledge transfer from base to novel classes. 
However, these methods often involve rigorous designs and resort to considerable parameters for better base adaptation (c.f. Tab.\ref{tab:parameter_comparison}), part of which will be further fine-tuned on novel classes.  
Moreover, they struggle to capture subtle differences between input images, resulting in worse performance on fine-grained benchmarks compared to coarse-grained ones. 
Consequently, an important question arises: \textit{can a lightweight yet effective strategy, with few parameters and low computational cost, learn domain-specific features for better adaptation and task-aware information, enabling training-free transfer?}

In this paper, we present a simple yet effective framework DSS-Prompt, which leverages pre-trained models through prompts in a novel synergistic way to enhance adaptivity and generalization ability of the embedding space. 
To be specific, it employs two types of prompts in each transformer layer: (1) static prompts, which help capture domain-specific features to first bridge the gap between pre-training and downstream data, and (2) dynamic prompts, which further grasp input-specific semantics and adaptively adjust their importance to acquire more generalizable prompts. 
Specially, to capture subtle differences between input images as much as possible, we leverage the pre-trained model BLIP~\cite{li2022blip} to acquire semantic knowledge from two different modalities and then generate the corresponding instance-aware dynamic prompts. This is inspired from the literature that more diverse prior knowledge of pre-training paradigms is beneficial for better few-shot learning~\cite{zhang2023prompt}. By doing so, the prompted embeddings are expected to grasp task-specific knowledge, thus enabling easy transfer from base to novel classes even without fine-tuning on few-shot novel classes. 

Our main contributions are summarized as follows. 
\begin{itemize}
    \item We propose a simple yet effective FSCIL framework that extends the pre-trained ViT with minimal modifications to form a strong FSCIL classifier. 
    
    \item We present a novel synergistic strategy that combines both static and dynamic prompts to capture domain-specific and task-specific knowledge, thus enabling both adaptivity and generalization ability of the embedding space. 
    
    \item We conduct extensive experiments on four benchmarks, including CIFAR100~\cite{krizhevsky2009learning}, CUB200~\cite{wah2011caltech}, miniImageNet~\cite{ravi2017optimization} and ImageNet-R~\cite{hendrycks2021many}, to show that on top of the prompted embeddings above, a simple prototype classifier can beat state-of-the-arts by a large margin, especially on the fine-grained benchmarks. 
\end{itemize}

\section{Related Work}
\label{sec:related_work}

\begin{figure*}[!htbp]
	\centering
	\includegraphics[width=0.9\linewidth]{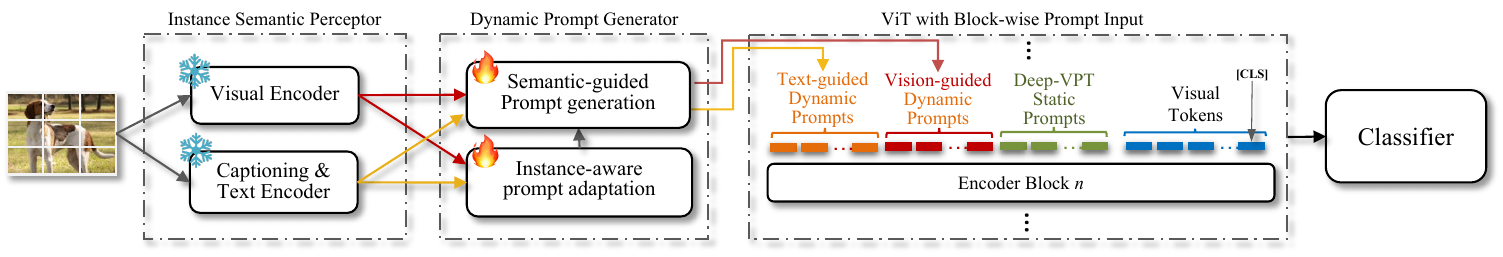}
    \caption{
    The schematic diagram of our DSS-Prompt model. For each input image, the \textbf{Instance Semantic Perceptor} (\textbf{ISP}) first extracts multi-modal prior knowledge, which is then fed into the \textbf{Dynamic Prompt Generator} (\textbf{DPG}) to produce instance-aware prompts. Finally, these dynamic prompts work in synergy with static prompts to get the final visual embeddings. 
    }
   \Description{The schematic design of Instance-Aware Dynamic Prompting model.}
    \label{fig:IDPromptDesign}
\end{figure*}

\textbf{Conventional Practice on CNNs for FSCIL.} As a newly proposed realistic setting, few-shot class-incremental learning (FSCIL) aims to continually learn new concepts from limited samples without catastrophically forgetting the previously learned ones at the same time and has attracted considerable attention since its inception. TOPIC~\cite{tao2020few} for the first time formulates the standard FSCIL setting and proposes to stabilize the feature topology among different sessions to alleviate the catastrophic forgetting issue. Follow-up methods could be roughly divided into four categories, including dynamic network-based methods~\cite{yang2021learnable, yang2022dynamic, gu2023few}, feature space-based methods~\cite{mazumder2021few, hersche2022constrained, yang2023neural}, meta learning-based methods~\cite{zhang2021few, zhou2022few, chi2022metafscil} and data replay-based methods~\cite{liu2022few, agarwal2022semantics, agarwal2022semantics}. For example, \cite{ahmed2024orco} proposes to learn orthogonal features in the representation space and reserve space for the future incremental data as the base samples during incremental learning are infeasible; \cite{zhao2023few} proposes a class-aware bilateral distillation method by adaptively drawing knowledge from two complementary teachers to reduce the overfitting in each novel session. These methods primarily use convolutional architectures (e.g., ResNet~\cite{he2016deep}) that are either learned from scratch or pre-trained on ImageNet~\cite{deng2009imagenet} to extract features and then exert various generalization strategies, which can only bring marginal performance improvements especially in novel sessions, given the limited model capacity. 

\textbf{Pre-trained Models for FSCIL.} Inspired by the recent remarkable successes in downstream tasks, the FSCIL community is exploring the powerful generalization capability of pre-trained ViT, either through purely visual comparison pre-training or vision-language contrastive pre-training, to achieve better knowledge transfer across successive sessions~\cite{d2023multimodal,doan2024streamlined,park2024pre,li2024few,goswami2024calibrating}. The basic idea is fixing the ViT backbone that has millions of parameters and only training a few new parameters through prompts or adapters. 
For example, \cite{d2023multimodal} first proposes to add prompts into both the vision and text encoders in CLIP~\cite{radford2021learning} and uses visual-textual feature comparison for prediction in FSCIL. While \cite{li2024few} further combines multi-modal prompts with feature adapters to reach a better feature alignment. In contrast, \cite{park2024pre} tunes the pre-trained knowledge through visual prompts and further distills semantic knowledge into the learning process to alleviate overfitting and catastrophic forgetting. \cite{liu2025few} introduces task-invariant and task-specific prompts between attention blocks in ViT and uses information bottleneck theory to alleviate the forgetting issue in old classes. \cite{wang2025approximation} thoroughly analyzes the approximation risk of FSCIL and presents practical guidelines when using ViT to maximize the classification margin across different sessions. By empirically comparing these methods, we found that even though ViT in CLIP is pre-trained on a broader image-text dataset (e.g., LAION-400M~\cite{schuhmann2021laion}) compared to ImageNet, its performance in FSCIL is slightly lower. This may be because the semantic gap between FSCIL benchmarks and the CLIP pre-training dataset is larger than the gap between FSCIL benchmarks and ImageNet. Similar to the latest work~\cite{park2024pre,liu2025few}, we build our approach on ViT pre-trained by ImageNet, combined with a prototype classifier in the embedding space for prediction.

\textbf{Prompt Learning for Vision Transformers.} 
Prompting learning is an effective learning scheme for vision transformers to adapt pre-trained knowledge to downstream tasks, without re-training the model parameters~\cite{zhou2022learning, zhou2022conditional, khattak2023maple, wang2022learning, smith2023coda, yao2023visual}. It generally appends learnable prompts to the input sequence in one or several layers and shows remarkable performance in retaining the model generalization ability and transferring to various downstream applications. For example, \cite{zhou2022learning, zhou2022conditional} propose to use learnable textual prompts to steer CLIP for low-shot classification, while \cite{khattak2023maple} proposes multi-modal prompting learning to obtain better alignment between the vision and language representations. \cite{yao2023visual} further reduces the discrepancy between the hand-crafted prompt and the trainable prompt token to alleviate forgetting. In class-incremental learning, \cite{wang2022learning} optimizes task-invariant and task-specific prompts to instruct the model to remember historical knowledge. \cite{smith2023coda} proposes to train the prompt pool and uses an end-to-end key-query scheme to improve model plasticity. Further, \cite{khan2023introducing} introduces language guidance at the task level in the prompt pool. However, all these methods need sufficient data samples to capture task-specific knowledge and thus are not suitable under the FSCIL setting. In contrast, we propose to generate input-aware prompts containing diverse multi-modal semantics, which have been proved to be beneficial to the few-shot task~\cite{zhang2023prompt}.

\section{Method}
\label{sec:method}

\subsection{Problem Definition}
Few-shot class-incremental learning (FSCIL) aims to develop a classification model capable of rapidly learning new visual concepts with few training samples while preserving the knowledge gained from previous classes. Formally, FSCIL consists of a base session, containing abundant labeled training data, and $T$ incremental sessions, each of which has only a few labeled data (e.g., 5-way 5-shot task) to update the model. Let $\mathcal{D} = \{\mathcal{D}^0, \mathcal{D}^1, ..., \mathcal{D}^T\}$ denote the training datasets for all sessions, where $\mathcal{D}^0$ and $\{\mathcal{D}^t\}_{t=1}^{T}$ are for the base session and $t$-th incremental session, respectively. In each session, we have $\mathcal{D}^t =  \{(\mathbf{x}^t_i, y^t_i)\}_{i=1}^{|\mathcal{D}^t|}$, where $\mathbf{x}^t_i$ denotes the $i$-th training image and $y^t_i$ denotes the corresponding label from the label space $\mathcal{Y}^t$. Between different sessions, there are no overlapped classes, i.e., $\forall j \neq t, \mathcal{Y}^j \cap \mathcal{Y}^t = \emptyset$. In this paper, we focus on the \textit{rehearsal-free} setting where no historical data can be fetched for replay. At $t$-th session, we can only access data from $\mathcal{D}^t$ for training and need to test the model on all previously seen classes $\mathcal{Y}^0 \cup \mathcal{Y}^1 \cup ... \cup \mathcal{Y}^t$.

\begin{figure*}[t]
    \centering
    \includegraphics[width=0.9\linewidth]{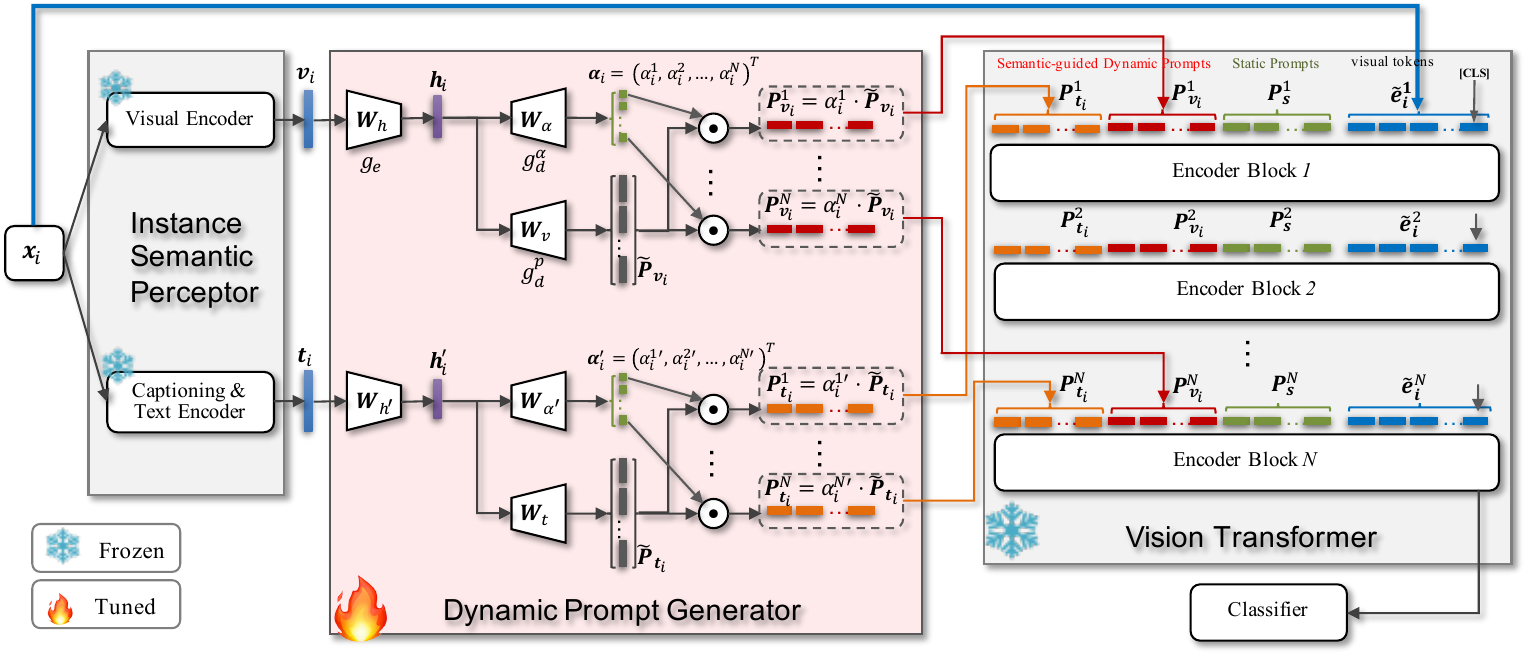}
    \caption{Details of our dynamic prompt generator. Given the multi-modal prior knowledge $\mathbf{h}_i$ (or $\mathbf{t}_i$) for an input $\mathbf{x}_i$, we use a encoder parameterized by $\mathbf{W}_h$ to map it to a low-dimensional latent space, and a two-headed decoder (parameterized by $\mathbf{W}_v$ and $\mathbf{W}_{\alpha}$) to generate the corresponding prompt tokens $\tilde{\mathbf{P}}_{v_i}$ and their scaling coefficients $\bm{\alpha}_i$. Then, considering each block differs in the level of semantics, we multiply $\tilde{\mathbf{P}}_{v_i}$ with $\alpha_i^n$ to get the adaptive dynamic prompts $\mathbf{P}_{v_i}^n$ for the $n$-th block. 
    }
    \Description{The details of the dynamic prompts.}
    \label{fig:IDPromptDetails}
\end{figure*}

Generally, a FSCIL model can be decoupled into a feature  backbone $f_{\theta}$ and a classifier $f_{\phi}$. We follow the current literature~\cite{wang2024few, liu2025few} that suggests: if strong generalized features could be learned in the base session, then a simple training-free prototype classifier is sufficient to achieve competitive performance. Specifically, with the embedding $f_{\theta}$ \textit{frozen} throughout the incremental learning process, we compute the average embedding (i.e., prototype) of each class as: 
\begin{equation}
    \mathbf{w}_c = \frac{1}{K} \sum_{i=1}^{|\mathcal{D}^t|} \mathbb{I}(y_i^t=c)f_{\theta}(\mathbf{x}_i^t), 
    \label{eq:prototype}
\end{equation}
where $K=\sum_{i=1}^{|\mathcal{D}^t|}\mathbb{I}(y_i^t=c)$ and $\mathbb{I}(\cdot)$ is the indicator function. The prototype represents the most common pattern within each class and can be used as the classifier $f_{\phi}$. Thus, for a test sample $\mathbf{x}$, the model predicts its label $y$ via the softmax probability with class prototypes: 
\begin{equation}
    p(y=c|\mathbf{x})= \frac{\text{exp} (\text{sim}(\mathbf{w}_c, f_{\theta}(\mathbf{x})))}{\sum_{j=1}^J \text{exp} (\text{sim}(\mathbf{w}_j, f_{\theta}(\mathbf{x})))}, 
    \label{eq:prediction}
\end{equation}
where $\text{sim}(\cdot)$ denotes the cosine similarity and $J=|\mathcal{L}^0| + |\mathcal{L}^1| + \ldots + |\mathcal{L}^t|$ is the total number of classes in $t$-th incremental session.

\subsection{Dynamic-Static Synergistic Prompting}
Pre-trained ViT has shown great generalization ability, which can be transferred to downstream tasks in a parameter-efficient fine-tuning manner through prompts. When used as the backbone network in FSCIL, on the one hand, it needs to better grasp the domain-specific knowledge, so as to bridge the distribution gap between the upstream pre-training dataset and the downstream dataset. On the other hand, it must maintain sufficient generalization ability, enabling easy knowledge transfer from base to novel classes, so as to avoid catastrophic forgetting. Therefore, this raises a question: with minimal parameters and computational resources, \textit{is there a simple yet effective strategy that can grasp domain-specific features for better adaptation and task-specific information for training-free transfer?} 

To this end, we introduce a novel approach termed Dynamic-Static Synergistic Prompting (DSS-Prompt) in Fig.\ref{fig:IDPromptDesign}, which adapts the pre-trained ViT using two types of prompts synergistically in each transformer block.
Specifically, it consists of: (1) static prompts to grasp domain-specific knowledge, which remain fixed in incremental sessions once learned; and (2) dynamic prompts to capture instance-specific semantic knowledge, which adaptively change according to the input in each session. 
In DSS-Prompt, we freeze the pre-trained ViT backbone and only optimize the small amount of trainable parameters in these prompts during the base session. The entire model remains frozen throughout the incremental learning process, enabling direct training-free prediction for FSCIL.

\textbf{Static Prompts for Domain-Specific Adaptation. } There exists a significant domain gap between the pre-training dataset and downstream data~\cite{zhou2024revisiting, liu2025few}. To bridge this gap, we adopt the deep prompting strategy~\cite{jia2022visual} that inserts prompts not only into the input layer but also into every later layer. More specifically, assume that a plain ViT has $N$ layers (i.e., blocks). An input image $\mathbf{x}_i$ is first split into $M$ equal-sized patches that are then transformed into sequence-like features $\mathbf{e}_i \in \mathbb{R}^{(M+1)\times d}$, where the last token in $\mathbf{e}_i$ denotes the ``[CLS]'' token to simplify notation. 
For each Transformer block that takes prompts, we define $L_s$ learnable visual prompts $\mathbf{P}_s^n \in \mathbb{R}^{L_s \times d}$, where $n$ denotes the $n$-th layer, and then append the prompts to form the extended features $\tilde{\mathbf{x}}_i^n$, i.e., 
\begin{equation}
    \tilde{\mathbf{x}}_i^n = [\mathbf{P}_s^n; \tilde{\mathbf{e}}_i^n]
    \label{eq:static_prompts}
\end{equation}
where $\tilde{\mathbf{e}}_i^n$ is the encoded features of $\mathbf{e}_i$ in the $n$-th layer. Empirically, learning only these static prompts in the base session could already help the model achieve promising FSCIL performance (c.f. Tab.\ref{tab:ablation_studies}). However, as pointed out in~\cite{zhou2022conditional}, if the prompts are optimized only for a specific set of classes and remain unchanged afterwards, the model tends to overfit these categories, leading to degraded transferability to wider novel classes. We therefore propose to further dynamically add input-aware information into prompts and adaptively adjust their importance in each layer.

\textbf{Dynamic Prompts for Task-Specific Transfer.} We use the following two steps to generate instance-aware prompts, as shown in Fig.\ref{fig:IDPromptDetails}: (1) extract as much semantic knowledge as possible from the input image using the proposed instance semantic perceptor; and (2) adaptively generate prompts using our dynamic prompt generator and append them to Eq.\ref{eq:static_prompts} to form the features.  

\textit{Step 1: Instance Semantic Perceptor (ISP) to Extract Complementary Information. } 
Diverse knowledge helps assist the representation learning, thereby facilitating the recognition of novel concepts~\cite{xing2019adaptive, zhang2023prompt}. Inspired by this, we suggest extracting both visual and semantic knowledge for each input image. We achieve this goal by simply leveraging a publicly available vision-language pre-trained model BLIP~\cite{li2022blip}. Specifically, for each input $\mathbf{x}_i$, we employ its image encoder to compute the visual knowledge $\mathbf{v}_i$. To extract the textual knowledge, we first use BLIP`s image-grounded text decoder to predict the textual descriptions, which are then encoded into embeddings $\mathbf{t}_i$ leveraging BLIP`s text encoder. $\mathbf{t}_i$ could be viewed as an additional perspective to supplement visual information. 
Notably, the ISP module is designed with flexibility in mind, enabling the integration of any effective semantic knowledge extractor to derive multi-modal embeddings for an image. 
This adaptability ensures compatibility with diverse and evolving approaches for semantic knowledge extraction. We leave exploration of more advanced designs for future work. 

\textit{Step 2: Dynamic Prompt Generator (DPG) to Produce Instance-aware Prompts. } Considering the requirement of parameter-efficient, we introduce a lightweight neural network to generate for each input image instance-aware tokens. Specifically, for the two types of knowledge extracted in Step 1, we generate the corresponding prompt tokens separately in the same manner, then append them to Eq.~\ref{eq:static_prompts} forming the input to each Transformer layer. For brevity, we only describe how to produce vision-guided dynamic prompts in detail. 

Let $g(\cdot)$ denote the generator network, which consists of a encoder $g_e(\cdot)$ mapping the input to a low-dimensional latent space and a two-headed decoder - one head $g_d^p(\cdot)$ for generating prompt tokens and the other $g_d^{\alpha}(\cdot)$ for adaptively generating layer-wise prompt scaling coefficients. Formally, we have: 
\begin{equation}
\begin{split}
    \mathbf{h}_i = g_e(\mathbf{v}_i|\mathbf{W}_h), \tilde{\mathbf{P}}_{v_i} = g_d^p(\mathbf{h}_i|\mathbf{W}_{v}), \bm{\alpha}_i = g_d^{\alpha}(\mathbf{h}_i|\mathbf{W}_{\alpha}), 
\end{split}
\end{equation}
where $\mathbf{h}_i \in \mathbb{R}^{d'}$, $\tilde{\mathbf{P}}_{v_i} \in \mathbb{R}^{L_{v} \times d}$ and $\bm{\alpha}_i \in \mathbb{R}^N$. 
In this work, $g_e$ is built with a fc-layer, reducing the input dimension by 3$\times$ (i.e., 768$\rightarrow$256). 
$g_d^p$ is a linear transform that maps the $d'$-dimensional latent representation $\mathbf{h}_i$ into a long output vector of $(L_{v}\times d)$ dimensions, which is then equally divided into $L_{v}$ prompt tokens of $d$ dimensions each.
$g_d^{\alpha}$ is built with a fc-layer, followed by the sigmoid activation function. 
$\mathbf{W}_h$, $\mathbf{W}_v$ and $\mathbf{W}_{\alpha}$ denotes the learnable parameters. Considering that each transformer layer differs in the level of semantics~\cite{dosovitskiy2020image,raghu2021vision,caron2020unsupervised}, we believe that each layer’s prompt should have a different weight based on the input when generating the final embedding.  Therefore, we get the vision-guided prompt for $n$-th transformer layer as: 
\begin{equation}
    \mathbf{P}_{v_i}^n = \alpha_i^n \tilde{\mathbf{P}}_{v_i}, 
    \label{eq:vision-guided-prompts}
\end{equation}
where $\alpha_i^n$ denotes the $n$-th element in $\bm{\alpha}_i$. Similarly, we can use the same way to generate the text-guided prompts $\mathbf{P}_{t_i}^n$. Note that when all elements of $\V{\alpha}_i$ are identical, each layer receives the same dynamic prompts; otherwise the prompts are adaptively adjusted based on the current input instance. Finally, we append these prompts to Eq.\ref{eq:static_prompts} to get the final input features for $n$-th Transformer layer as: $ \tilde{\mathbf{x}}_i^n = [\mathbf{P}^n_{t_i}; \mathbf{P}^n_{v_i}; \mathbf{P}_s^n; \tilde{\mathbf{e}}_i^n]$. 

\begin{figure}[t]
    \centering
    \includegraphics[width=0.9\linewidth]{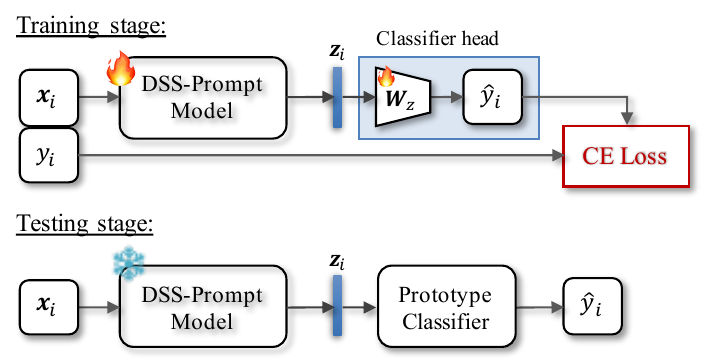}
    \caption{Illustration of the training procedure. On top of the prompted features, we optimize the linear classifier with the standard CE loss in the training phase and employ the simple prototype classifier for predicting class labels in the testing phase.  }
    \Description{The details of the classifier}
    \label{fig:Classifier}
\end{figure}

\textbf{Training Procedure.} As shown in Fig.\ref{fig:Classifier}, during training in the base session, we compute the prompted feature embeddings $\mathbf{z}_i$ for each input image $\mathbf{x}_i$, i.e., the output embedding corresponding to the ``[CLS]'' token, and use the linear classifier with parameters $\mathbf{W}_z$ for prediction. We use the standard cross-entropy loss to optimize the small amount of learnable parameters, including $\mathbf{W}_h$, $\mathbf{W}_{h'}$, $\mathbf{W}_v$, $\mathbf{W}_{\alpha}$, $\mathbf{W}_t$, $\mathbf{W}_{\alpha'}$, $\mathbf{W}_z$, $\{\mathbf{P}_s^n\}_{n=1}^N$, while keeping the backbone fixed. For testing, we use the prototype classifier following~\cite{zhou2022forward,wang2024few,liu2025few}.
The whole FSCIL procedure is described in Alg.\ref{alg:IDPrompt}. Since the model is frozen in the subsequent tasks, it does not suffer catastrophic forgetting of base concepts while dynamically adjusting to novel concepts. 

\begin{algorithm}
\caption{Dynamic-Static Synergistic Prompting (DSS-Prompt) for FSCIL}
\begin{algorithmic}[1]
    \REQUIRE Incremental datasets $\{\mathcal{D}^0, \mathcal{D}^1, ..., \mathcal{D}^T\}$, Pre-trained ViT; 
    \ENSURE Updated model;
    \STATE Adapt ViT through dynamic-static prompts to $\mathcal{D}^0$ via minimizing the cross-entropy loss;
    \STATE Freeze the whole model; 
    \FOR{$t \gets 1$ to $T$}
        \STATE Extract visual embeddings for $\mathcal{D}^t$; 
        \STATE Compute class prototypes via Eq.~\ref{eq:prototype};
        \STATE Use all prototypes to predict class labels via Eq.\ref{eq:prediction}; 
    \ENDFOR
\end{algorithmic}
\label{alg:IDPrompt} 
\end{algorithm}

\begin{table*}[t!]
	\centering
	\resizebox{0.95\textwidth}{!}{
		\begin{tabular}{c c c c c c c c c c c c c c}
			\hline
			\multirow{2}{*}{Methods} &  \multicolumn{11}{c}{Accuracy in each session (\%) $\uparrow$} & \multirow{2}{*}{AVG$\uparrow$} & \multirow{2}{*}{PD$\downarrow$}\\
			\cmidrule{2-12}
			& 0 & 1 & 2 & 3 & 4 & 5 & 6 & 7 & 8 & 9 & 10 & & \\
			\midrule
            CLIP-M$^{\text{3}}$~\cite{doan2024streamlined} & 84.50 & 81.90 & 80.70 & 78.40 & 77.80 & 77.00 & 76.10 & 76.00 & 74.80 & 75.10 & 74.90 & 77.93 & 9.60 \\
            LP-DiF~\cite{huang2024learning}  & 83.94 & 80.59 & 79.17 & 74.30 & 73.89 & 73.44 & 71.60 & 70.81 & 69.08 & 68.74 & 68.53 & 74.01 & 15.41 \\ 
            CPE-CLIP~\cite{d2023multimodal}& 81.58 & 78.52 & 76.68 & 71.86 & 71.52 & 70.23 & 67.66 & 66.52 & 65.09 & 64.47 & 64.60 & 70.79 & 16.98 \\ 
            CMA-FSCIL~\cite{CMA-FSCIL}   & 84.78 & 80.50 & 79.21 & 74.46 & 74.65 & 74.23 & 72.65 & 70.33 & 69.84 & 70.33 & 70.40 & 74.70 & 14.38 \\ 
            CA-CLIP~\cite{xu2024clip} & 85.16 & 79.20 & 77.49 & 70.97 & 70.82 & 69.12 & 66.33 & 62.54 & 60.55 & 61.05 & 60.33 & 69.41 & 24.83 \\ 
            \midrule
			CEC~\cite{zhang2021few} & 84.80 & 82.50 & 81.40 & 78.50 & 79.30 & 77.80 & 77.40 & 77.60 & 77.20 & 76.90 & 76.80 & 79.11 & 8.00 \\ 
			FACT~\cite{zhou2022forward} & 87.30 & 84.20 & 82.10 & 78.10 & 78.40 & 76.30 & 75.40 & 75.50 & 74.40 & 74.10 & 73.90 & 78.15 & 13.40 \\ 
			TEEN~\cite{wang2024few}  & 89.00 & 86.50 & 85.90 & 83.30 & 83.30 & 82.20 & 82.10 & 80.40 & 80.50 & 80.10 & 80.60 & 83.08 & 8.40 \\ 
			PL-FSCIL~\cite{tian2024pl} & 85.16 & 85.40 & 82.75 & 75.22 & 77.22 & 73.25 & 72.39 & 70.24 & 67.97 & 68.33 & 69.86 & 75.25 & 15.30 \\ 
			SV-T~\cite{qiu2023semantic} & 84.19 & 82.63 & 81.21 & 78.97 & 79.38 & 77.64 & 77.55 & 75.71 & 75.91 & 75.77 & 76.17 & 78.65 & 8.02 \\ 
			AR-FSCIL~\cite{wang2025approximation}  & 86.69 & 85.38 & 84.34 & 82.43 & 83.08 & 81.07 & 81.38 & 81.42 & 81.21 & 81.11 & 81.30 & 82.67 & 5.39 \\ 
			PriViLege~\cite{park2024pre} & 82.21 & 81.25 & 80.45 & 77.76 & 77.78 & 75.95 & 75.69 & 76.00 & 75.19 & 75.20 & 75.08 & 77.51 & 7.13 \\ 
			ASP~\cite{liu2025few} & 87.10 & 86.00 & 84.90 & 83.40 & 83.60 & 82.40 & 82.60 & 83.00 & 82.60 & 83.00 & 83.50 & 83.83 & \textbf{3.60} \\ 
			\textbf{DSS-Prompt} & \textbf{90.95} & \textbf{87.64} & \textbf{86.43} & \textbf{85.12} & \textbf{85.32} & \textbf{83.99} & \textbf{83.59} & \textbf{83.63} & \textbf{83.25} & \textbf{83.62} & \textbf{84.22} &\textbf{85.25} & 6.73 \\
			\hline
        \end{tabular}
    }
    \caption{FSCIL performance comparison on CUB200 with ViT-B/16 as the backbone. ``AVG'' represents the average Top-1 accuracy of all sessions; ``PD'' represents the performance drop rate. $\uparrow$/$\downarrow$ means higher/lower is better. The best performance is in \textbf{bold}.  }
    \label{tab:sota-cub200}
\end{table*}

\section{Experiments}
\subsection{Experimental Setup}
\textbf{Datasets. }
Following the standard practice in FSCIL~\cite{tao2020few,liu2025few}, we evaluate our method on four benchmarks: CIFAR100~\cite{krizhevsky2009learning}, miniImageNet~\cite{ravi2017optimization}, CUB200~\cite{wah2011caltech}, and ImageNet-R~\cite{hendrycks2021many}. Tab.~\ref{tab:dataset_statistics} lists the detailed split configuration in all datasets. 

\textbf{Metrics. } 
We evaluate the model performance with the test set and primarily report the average Top-1 accuracy (denoted as $\mathcal{A}_t$) in each session $t$ and the average accuracy (denoted as ``AVG'') across all sessions. To measure the forgetting issue, we also report the performance dropping rate (denoted as ``PD'') that measures the absolute accuracy drop in the last session $\mathcal{A}_T$ w.r.t. the accuracy in the base session $\mathcal{A}_0$, i.e., $\text{PD} = \mathcal{A}_0 - \mathcal{A}_T$. 

\begin{table}[t!]
    \centering
    \small
    \begin{tabular}{c c c c c }
    \toprule
    \textbf{Dataset} & \textbf{Sessions} & \textbf{Base} & \textbf{Novel} & \textbf{Incremental}\\ 
     \midrule
     CIFAR100 & 1+8 & 60 & 40 & 5-way 5-shot \\
     CUB200 & 1+10 & 100 & 100 & 10-way 5-shot \\
     miniImageNet & 1+8 & 60 & 40 & 5-way 5-shot  \\
     ImageNet-R & 1+10 & 100 & 100 & 10-way 5-shot\\
     \bottomrule
    \end{tabular}
    \caption{Configuration settings for FSCIL benchmarks. ``Base'' and ``Novel'' denotes the number of base and novel classes respectively. ``Incremental'' denotes the task configuration in incremental sessions. }
    \vspace{-6mm}
    \label{tab:dataset_statistics}
\end{table}

\begin{figure*}[t!]
    \centering
    \begin{subfigure}[b]{0.33\linewidth}
        \centering
        \includegraphics[width=\textwidth]{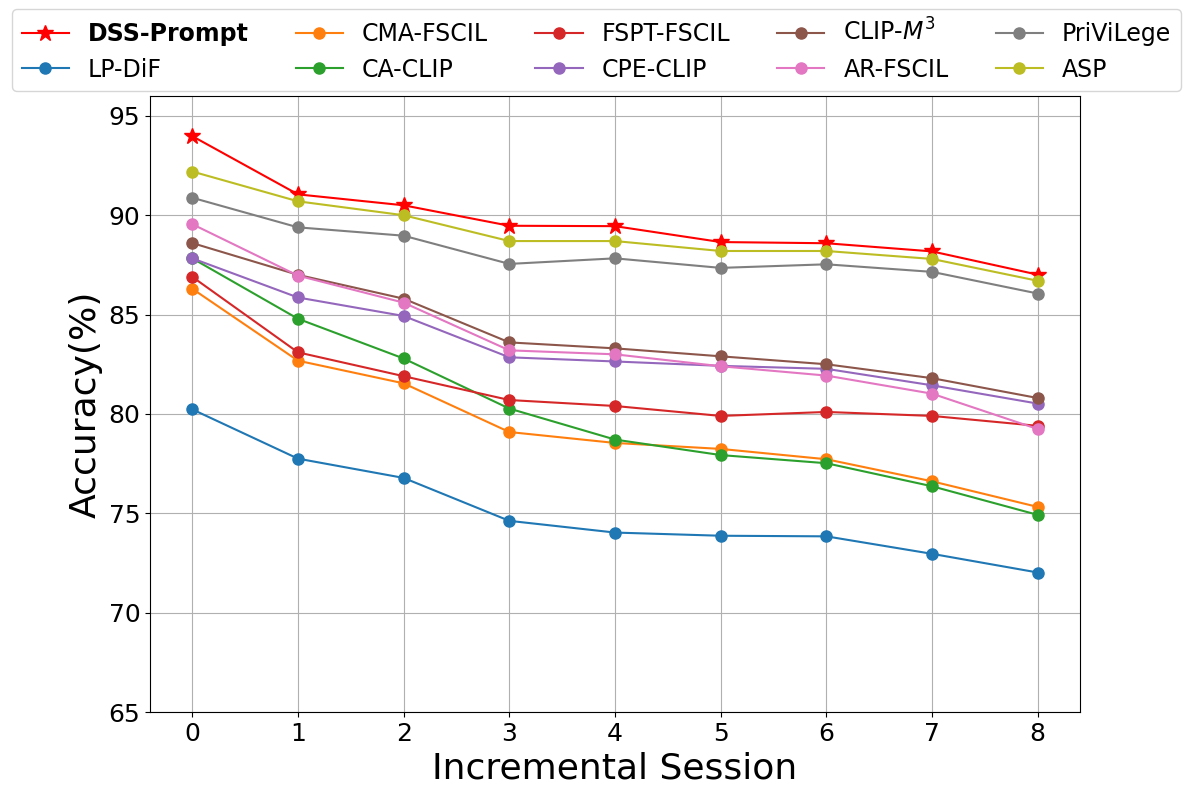}
    \end{subfigure}
    \hfill
    \begin{subfigure}[b]{0.33\textwidth}
        \centering
        \includegraphics[width=\textwidth]{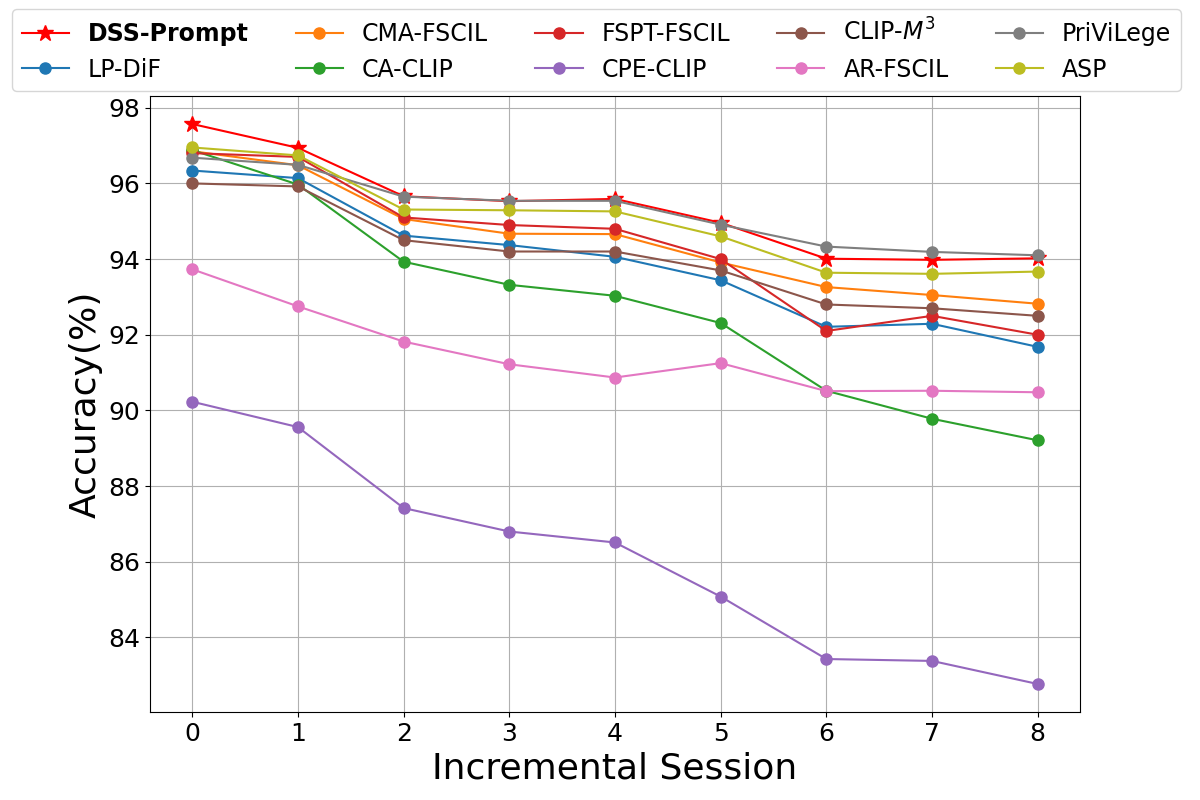}
    \end{subfigure}
    \hfill
    \begin{subfigure}[b]{0.33\textwidth}
        \centering
        \includegraphics[width=\textwidth]{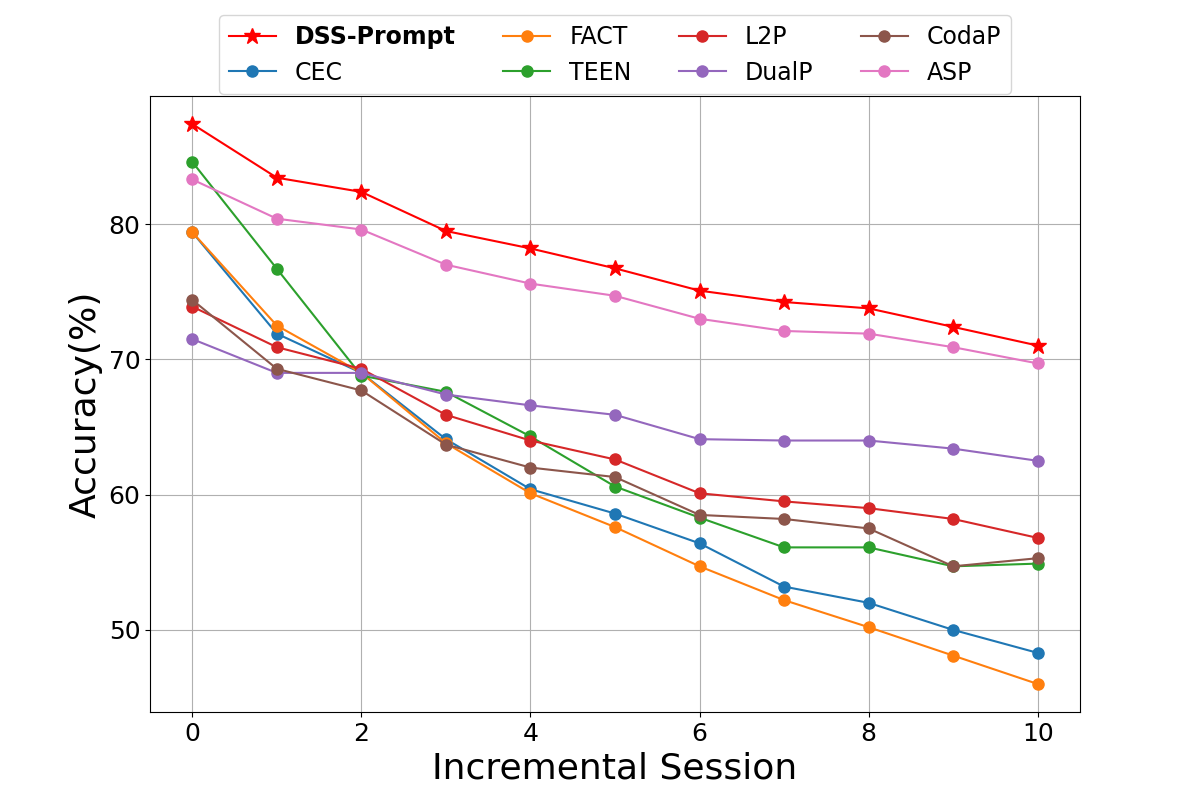}
    \end{subfigure}
    \caption{FSCIL performance comparison with ViT-B/16 as the backbone on CIFAR100 (left), miniImageNet (middle) and ImageNet-R (right). DSS-Prompt outperforms others in almost all sessions and exhibits less forgetting.}
    \Description{FSCIL performance comparison}
    \label{fig:sota-cifar-imagenet}
\end{figure*}

\textbf{Implementation Details.} We conduct all experiments with PyTorch~\cite{paszke2019pytorch} on two NVIDIA RTX 3080 Ti GPUs. Following~\cite{liu2025few}, we use ViT-B/16 pre-trained on ImageNet as the backbone and resize the input images to $224\times 224$. In the base session, we train our model using the SGD optimizer with the learning rate 0.01, momentum 0.9 and weight decay 0.03 for 60 epochs on CUB200 and 40 epochs for the others. During training, we use batch size of 24 for both CUB and ImageNet-R, and batch size of 16 for both CIFAR100 and miniImageNet. We set the same length for each type of prompts, i.e. $L_s = L_{v}$, with the number of tokens being 10 on ImageNet-R and 3 on others.

\begin{table*}[t]
    \centering
    \resizebox{\textwidth}{!}{
    \begin{tabular}{c c c c c c c c c c c c c c c c}
    \toprule
    \multirow{2}{*}{SP} & \multirow{2}{*}{DP} & \multirow{2}{*}{DP$_{\alpha}$} & \multicolumn{11}{c}{Accuracy in each session (\%) $\uparrow$} & \multirow{2}{*}{AVG $\uparrow$}  & \multirow{2}{*}{PD $\downarrow$} \\
    \cmidrule{4-14}
    & & & 0 & 1 & 2 & 3 & 4 & 5 & 6 & 7 & 8 &9 &10 & & \\
    \midrule
    $\times$ & $\times$ & $\times$ & 86.68 & 85.20 & 83.78 & 82.52 & 82.39 & 80.16 & 80.04 & 80.17 & 79.79 & 80.04 & 80.36 & 81.92 & 6.32 \\
    $\checkmark$ & $\times$ & $\times$  &  87.11 & 86.14 & 84.78 & 83.65 & 83.73 & 82.16 & 81.98 & 82.18 & 81.73 &82.15 &82.53 & 83.47 & 4.58 \\
    $\checkmark$ & $\checkmark_{v}$ & $\times$ & 87.96 & 86.54 & 85.00 & 83.79 & 83.91 & 82.33 & 82.25 & 82.53 & 82.06 &82.46 &82.87 & 83.79 & 5.09 \\
    $\checkmark$ & $\checkmark_{v}$ & $\checkmark$ & 87.87 & 86.69 & 85.21 & 84.05 & 84.28 & 82.79 & 82.73 & 82.83 & 82.30 &82.73 & 83.21& 84.06  & 4.66 \\
    $\checkmark$ & $\checkmark_{t}$ & $\times$ & 87.53 & 86.22 & 84.71 & 83.32 & 83.49 & 82.28 & 82.19 & 82.48 & 82.11 &82.60 &83.12 & 83.64 & \textbf{4.41} \\
    $\checkmark$ & $\checkmark_{t}$ & $\checkmark$ & 88.13 &86.77 &85.14 &83.85 &84.10 &82.68 &82.68 &82.88 &82.54 &83.00 &83.42 &84.11 &4.71 \\
    $\checkmark$ & $\checkmark_{v+t}$ & $\times$ & 88.30 & 87.09 & 85.64 & 84.45 & 84.53 & 83.19 & 83.22 & 83.23 & 82.87&83.27 &83.63 & 84.49  & 4.67 \\
    $\checkmark$ & $\checkmark_{v+t}$ &  $\checkmark$ & \textbf{90.95} & \textbf{87.64} & \textbf{86.43} & \textbf{85.12} & \textbf{85.32} & \textbf{83.99} & \textbf{83.59} & \textbf{83.63} & \textbf{83.25} &\textbf{83.62} &\textbf{84.22 }& \textbf{85.25 } & 6.73 \\
    \bottomrule
    \end{tabular}
    }
    \caption{Ablation studies on each component in our DSS-Prompt on CUB200. ``SP'' and ``DP'' denote static and dynamic prompts, respectively. DP without subscript means no scaling coefficients are used, i.e., $\alpha_i^n=1$, while with $\alpha$ means using adaptive scaling coefficients. 
    ``v'' and ``t'' in DP stand for vision- and textual-guided prompts, respectively. 
    The first row corresponds to the baseline that takes the pre-trained ViT with a prototype classifier. Each component contributes to performance improvement.}
    \label{tab:ablation_studies}
\end{table*}

\subsection{Comparison with State-of-the-Arts}
We compare our DSS-Prompt with other approaches on these benchmarks in Tab.\ref{tab:sota-cub200} and Fig.\ref{fig:sota-cifar-imagenet}, where all methods are based on the pre-trained ViT-B/16. 
From these results, we make four observations. (1) Our proposed method DSS-Prompt outperforms all competitors with a considerable margin on all datasets, especially in terms of the accuracy in each session and ``AVG''. Specifically, it surpasses ASP by 1.42\% AVG on CUB200. 
(2) Approaches that utilize multi-modal comparison for prediction based on CLIP (i.e. methods in the top part of Tab.\ref{tab:sota-cub200}) perform slightly worse than those utilizing pure visual comparison (i.e. methods in the bottom part of Tab.\ref{tab:sota-cub200}) for FSCIL, in terms of both AVG and PD. For example, CLIP-M$^{\text{3}}$ is clearly inferior to ASP. This may be because the pre-training dataset ImageNet exhibits a smaller distribution gap with FSCIL benchmarks than LAION-400M in CLIP, thus leading to better performance. 
(3) By comparing Tab.\ref{tab:sota-cub200} with Fig.\ref{fig:sota-cifar-imagenet}, DSS-Prompt shows a greater improvement on the fine-grained dataset. Specifically, it surpasses ASP by 1.42\% AVG on CUB200, while 0.64\% AVG on CIFAR100. Given the fewer samples per class (e.g., 60 in CUB200 v.s. 500 in CIFAR100), acquiring generalized knowledge in the base session and then transferring it to the incremental sessions becomes more challenging. Our instance-aware prompts clearly can help capturing more effective knowledge. (4) Comparing with most methods, DSS-Prompt can effectively mitigate the catastrophic forgetting. However, it ranks only in 2nd or 3rd in the PD performance, which shows a possible direction for our future work.

\subsection{Ablation Study}
\textbf{Effectiveness of each component. } 
We conduct a series of ablation studies on CUB200 to validate each component in our DSS-Prompt.  
We take the pre-trained ViT with a prototype classifier as the baseline, to which we sequentially add each component in Tab.~\ref{tab:ablation_studies}. 
We notice that: (1) Deep static prompts are effective in adapting pre-trained knowledge to downstream FSCIL tasks. (2) Our instance-aware prompts further improve the performance, validating the necessity of instance-specific information for improving generalization to novel classes. (3) Multi-modal prompts in our dynamic prompts may provide complementary semantics, thereby boosting performance further. For example, using ''v+t'' achieves 84.49\%, better than 83.79\% or 83.64\% when using only ''v'' or ''t''. 
(4) Applying different weights to the generated prompts in each layer of ViT is more beneficial, e.g., 85.05\% vs 84.49\% in AVG. This is understandable, since each layer captures different semantic information. 
In conclusion, our DSS-Prompt demonstrates outstanding performance with significant margins compared to the baseline. 

\begin{figure}[htbp]
\centering
    \begin{subfigure}[b]{0.23\textwidth}
        \centering
        \includegraphics[width=\textwidth]{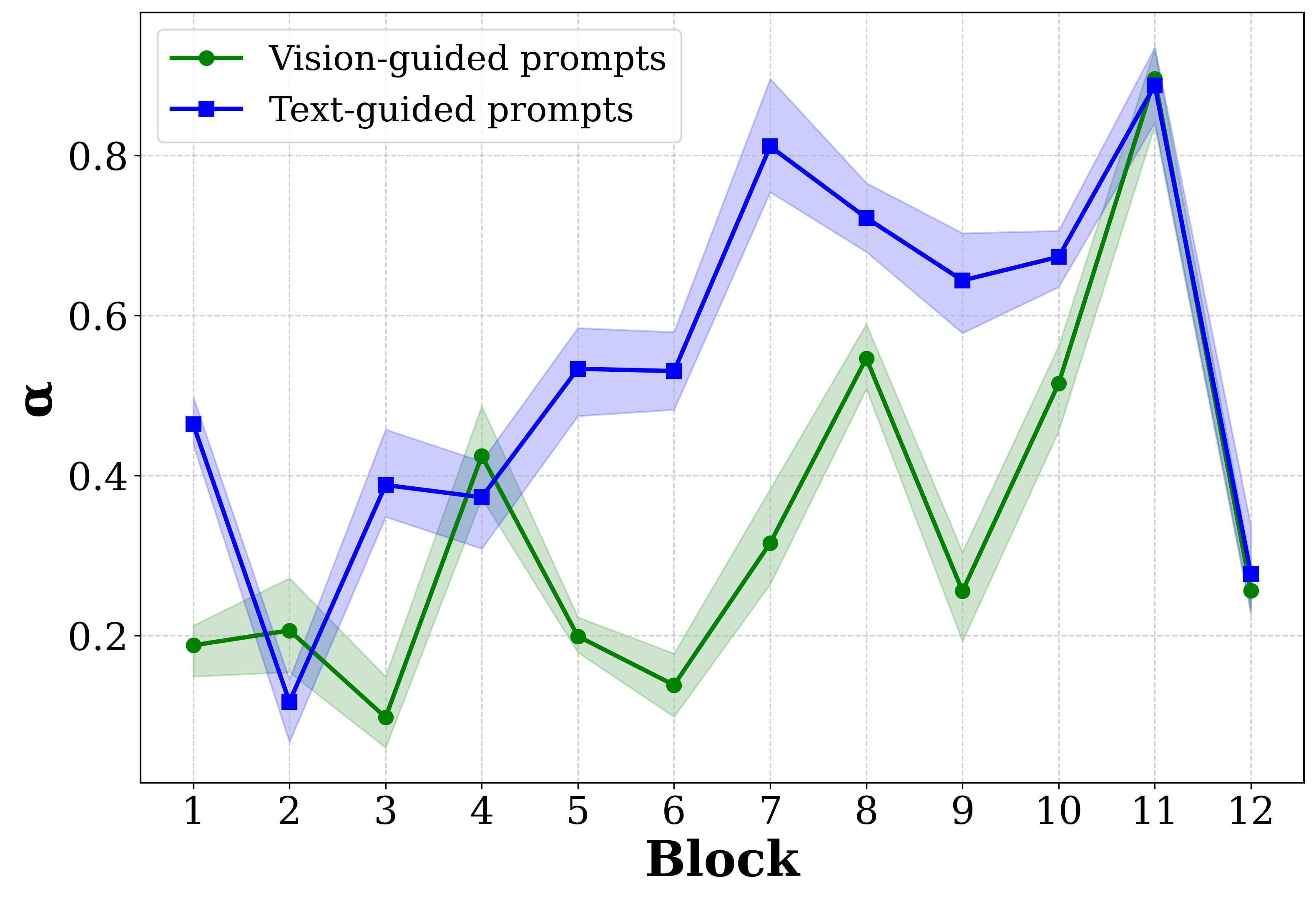}
    \end{subfigure}
    \hfill
    \begin{subfigure}[b]{0.23\textwidth}
        \centering
        \includegraphics[width=\textwidth]{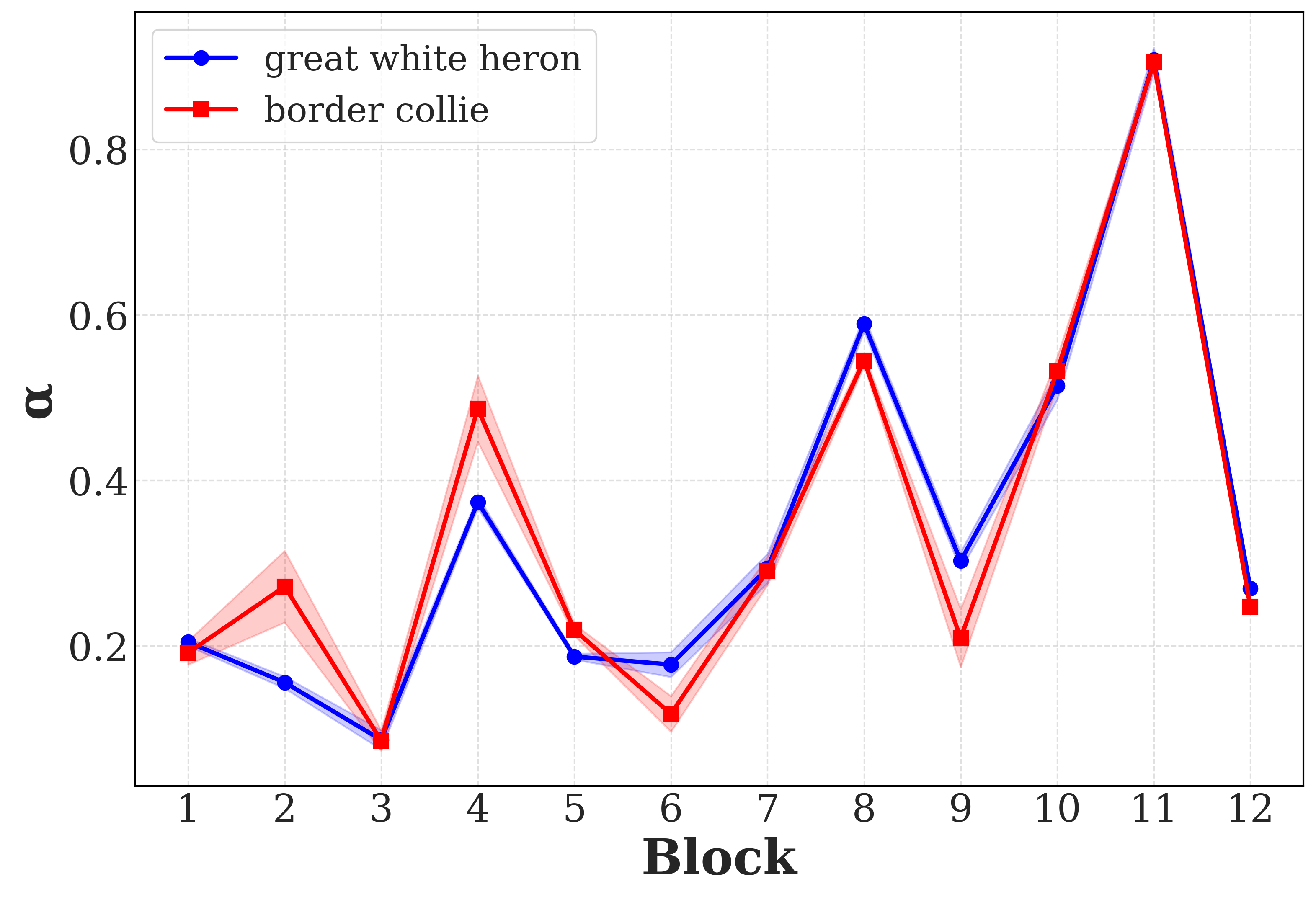}
    \end{subfigure}
    \caption{Left: the scaling coefficient distribution across all samples in each ViT block on ImageNet-R. Right: the scaling coefficients for visual-guided prompts of two random classes in ImageNet-R. }
    \Description{Ablation studies on CUB.}
    \label{fig:coefficients_distribution}
\end{figure}

\begin{figure*}[!htp]
    \centering
    \includegraphics[width=0.92\textwidth]{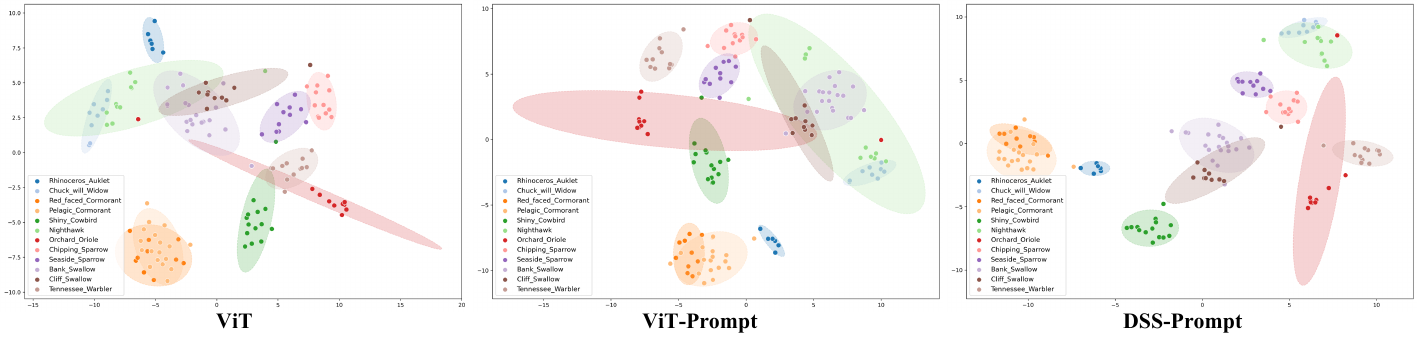}
    \caption{T-SNE Visualizations of feature embeddings on CUB200, obtained by the pre-trained ViT, the ViT-Prompt baseline that inserts static prompts into each transformer layer, and our DSS-Prompt. Our method achieves a clearer decision boundary. }
    \Description{TSNE visualizations.}
    \label{fig:tsne-visualization}
\end{figure*}

\textbf{Distribution of scaling coefficients in dynamic prompts (DP). } Our DSS-Prompt adaptively determines the importance of instance-aware prompts in each layer via the scaling coefficients in DP, instead of using the same weight 1.  
For further analysis, Fig.\ref{fig:coefficients_distribution} (left) shows the scaling coefficient distribution across each ViT block on ImageNet-R, while Fig.\ref{fig:coefficients_distribution} (right) displays the coefficient distribution for visual-guided prompts of two random classes. 
We can see that: (1) The coefficients differ in different modalities, providing complementary information in DP. (2) The coefficients vary across different categories as well, indicating that DP may capture class-specific knowledge to boost discrimination. 

\begin{figure}[!htp]
   \includegraphics[width=0.4\textwidth]{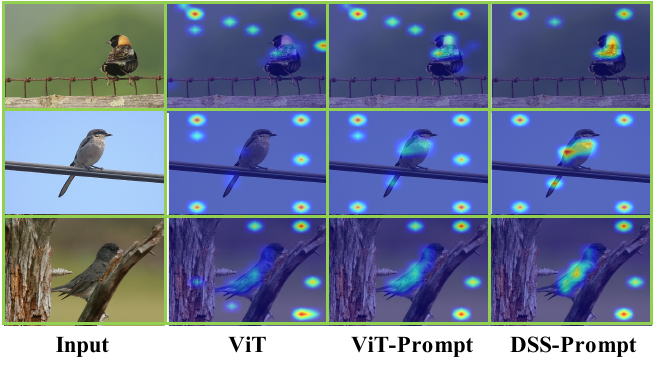}
    \caption{Attention Rollout visualizations of the pre-trained ViT, the baseline ViT-Prompt and our DSS-Prompt. Important regions are shown with warm colors. DSS-Prompt concentrates more on object-specific regions. }
    \Description{fig:attention-rollout-cub}
    \label{fig:attention-rollout-cub}
\end{figure}

\textbf{Visualization of feature embeddings.} We visualize the feature embeddings, obtained from the pre-trained ViT, the baseline ViT-Prompt that adds static prompts into every transformer block, and our DSS-Prompt, with t-SNE~\cite{van2008visualizing} on CUB200 dataset in Fig.\ref{fig:tsne-visualization}. As shown in this figure, some overlapped classes become more compact and exhibit clearer decision boundaries, verifying the adaptivity and generalizability of DSS-Prompt. We also visualize the Attention Rollout~\cite{abnar2020quantifying} results to highlight the critical regions in the image for predicting the corresponding concept in Fig.~\ref{fig:attention-rollout-cub}. DSS-Prompt concentrates more on class-specific features than vanilla ViT.

\textbf{Balanced generalization across base and novel classes. } In addition to the average Top-1 accuracy, we further report the harmonic mean accuracy~\cite{peng2022few,ahmed2024orco} between base and novel classes after the final session in Tab.~\ref{tab:harmonic_mean}.  
We see that our DSS-Prompt demonstrates superior performance across all three datasets, surpassing previous SOTA ASP by a promising margin. 

\begin{table}[!htp]
    \centering
    \begin{tabular}{|c|c|c |c|}
    \hline
        Methods & CIFAR100 & CUB200 & ImageNet-R  \\
    \hline
        CEC~\cite{zhang2021few} &  64.1 & 76.2 & 32.6 \\
        FACT~\cite{zhou2022forward} & 55.5 & 72.0 &22.3\\
        TEEN~\cite{wang2024few}  & 81.2 & 80.2 & 45.4\\
        ASP~\cite{liu2025few} & 85.3 & 83.4 & 67.0\\
        \textbf{DSS-Prompt} & \textbf{85.6 }& \textbf{84.0} & \textbf{67.5}\\
    \hline
    \end{tabular}
    \caption{The harmonic mean accuracy between base and novel classes after the final session. Following ASP, we compare DSS-Prompt with only those reporting detailed results on three benchmarks. }
    \label{tab:harmonic_mean}
    \vspace{-4mm}
\end{table}

\begin{figure}[htbp]
\centering
    \begin{subfigure}[b]{0.23\textwidth}
        \centering
        \includegraphics[width=\textwidth]{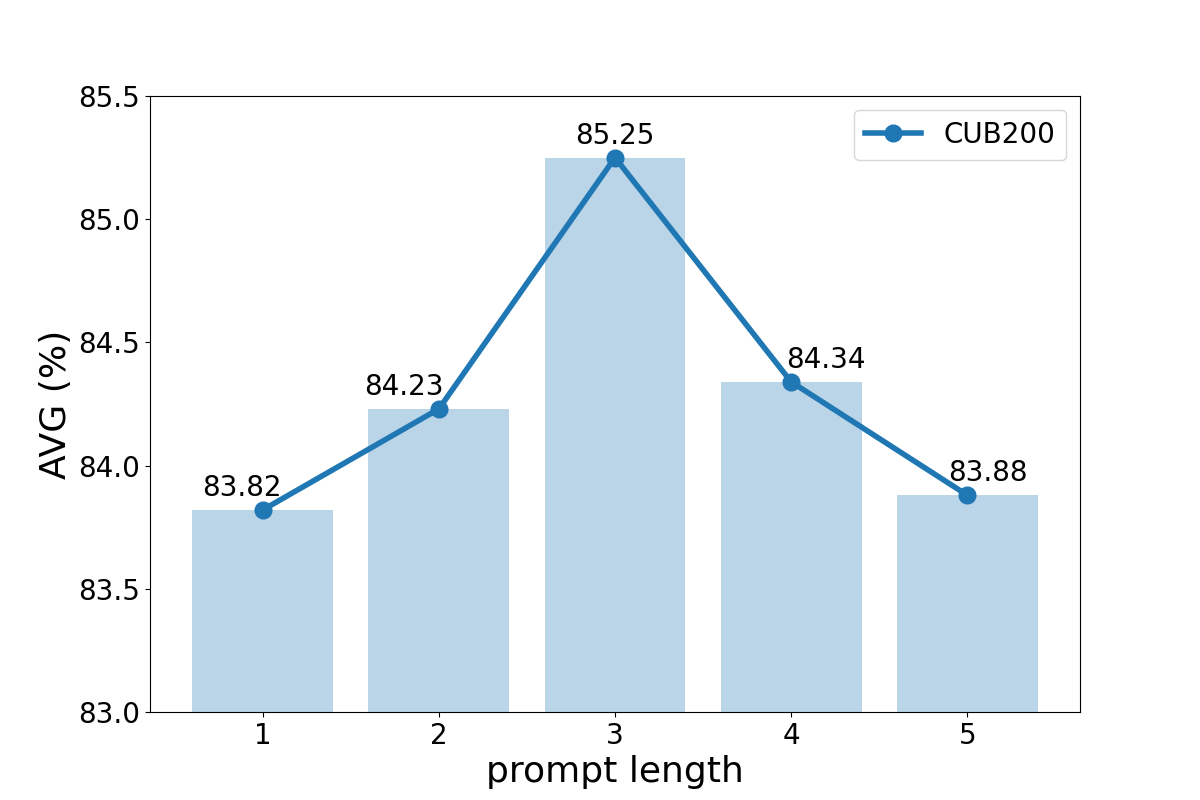}
    \end{subfigure}
    \hfill
    \begin{subfigure}[b]{0.23\textwidth}
        \centering
        \includegraphics[width=\textwidth]{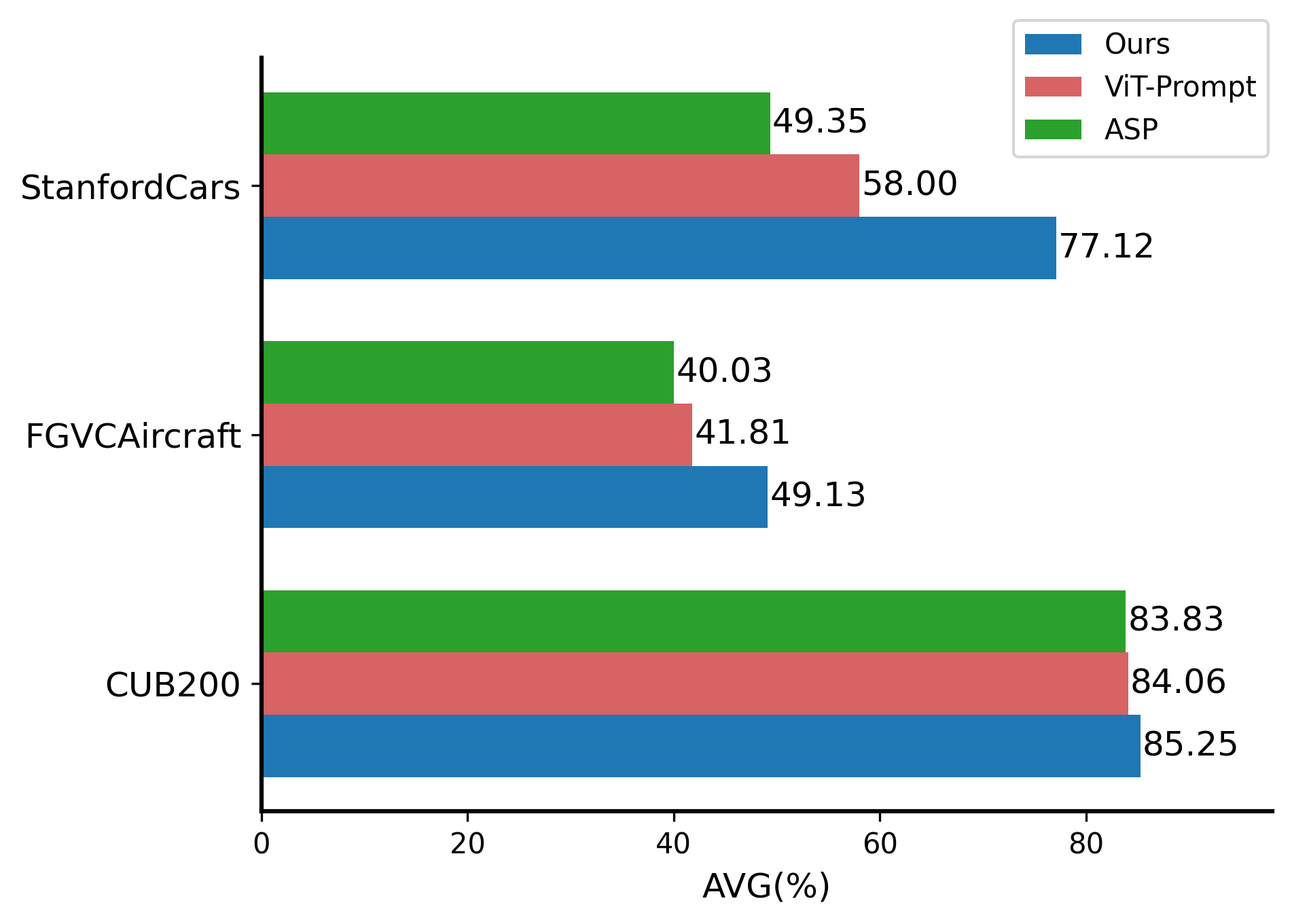}
    \end{subfigure}
    \caption{Left: ablation on prompt length on CUB200. Right: Performance comparison with ViT-Prompt and ASP~\cite{liu2025few} on fine-grained datasets. Our dynamic prompts yield higher performance gain.}
    \Description{fig:prompt_length_cub200}
	\label{fig:prompt_length_fine_grained}
\end{figure}

\textbf{Prompt length. } We study the impact of prompt length on the final FSCIL performance on CUB200 dataset by setting it to $\{1, 2, 3, 4, 5\}$ respectively. Fig.~\ref{fig:prompt_length_fine_grained} (left) displays the corresponding AVG performance. We can observe that as the prompt length increases, the performance gradually increases until it reaches the highest value in $3$. This is because too many prompts would overfit to base classes, which inherently hurts the generalization to novel classes when the training data is scarce.

\textbf{Better discrimination on fine-grained datasets.} Comparing the performance on CUB200 in Tab.\ref{tab:sota-cub200} and other FSCIL benchmarks in Fig.\ref{fig:sota-cifar-imagenet}, our method achieves greater improvement in fine-grained classification. We speculate that it can be attributed to the usage of input-conditioned dynamic prompts in each layer. To contextualize this, we compare with ViT-Prompt and ASP on other two fine-grained datasets: FGVCAircraft~\cite{maji2013fine} and StanfordCars~\cite{krause20133d} under the FSCIL setting. From Fig.\ref{fig:prompt_length_fine_grained} (right), we observe that employing dynamic prompts yields higher performance gain. 
We leave exploration on more datasets for future work.

\textbf{Parameter number. } Tab.\ref{tab:parameter_comparison} illustrates the parameter comparison among different prompting methods. Ours has fewer parameters while achieving higher performance.

\begin{table}[!htp]
    \centering
    \begin{tabular}{|c|c|c|c| }
    \hline
         Method & PriViLege~\cite{park2024pre} & ASP~\cite{liu2025few} & \textbf{DSS-Prompt} \\
    \hline
    \# Params. & 85.95M & 2.00M & 1.61M \\
    \hline
    \end{tabular}
    \caption{Parameter comparison with two closely related SOTA methods, i.e., PriViLege and ASP, for FSCIL. }
    \vspace{-7mm}
    \label{tab:parameter_comparison}
\end{table}

\section{Conclusion}
We present DSS-Prompt, a simple yet effective few-shot class-incremental learning approach that enables the pre-trained Vision Transformer to incrementally learn new concepts without catastrophically forgetting the old ones at the same time in a parameter-efficient manner. Specifically, it exerts deep static prompts and instance-aware multi-modal dynamic promtps into each layer of the architecture and adaptively adjusts their importance for the final prediction.
Extensive experiments show that incorporated with a simple prototype classifier, these prompted features outperform state-of-the-art methods with clear margins. In terms of future work, one direction is to further alleviate the forgetting and reach a well balance between base and novel sessions.

\begin{acks}
This work is supported by the National Natural Science Foundation of China (NSFC) under Grant No.62206256. 
\end{acks}

\bibliographystyle{ACM-Reference-Format}
\bibliography{acmmm-ref}


\begin{thebibliography}{63}


\ifx \showCODEN    \undefined \def \showCODEN     #1{\unskip}     \fi
\ifx \showISBNx    \undefined \def \showISBNx     #1{\unskip}     \fi
\ifx \showISBNxiii \undefined \def \showISBNxiii  #1{\unskip}     \fi
\ifx \showISSN     \undefined \def \showISSN      #1{\unskip}     \fi
\ifx \showLCCN     \undefined \def \showLCCN      #1{\unskip}     \fi
\ifx \shownote     \undefined \def \shownote      #1{#1}          \fi
\ifx \showarticletitle \undefined \def \showarticletitle #1{#1}   \fi
\ifx \showURL      \undefined \def \showURL       {\relax}        \fi
\providecommand\bibfield[2]{#2}
\providecommand\bibinfo[2]{#2}
\providecommand\natexlab[1]{#1}
\providecommand\showeprint[2][]{arXiv:#2}

\bibitem[Abnar and Zuidema(2020)]%
        {abnar2020quantifying}
\bibfield{author}{\bibinfo{person}{Samira Abnar} {and} \bibinfo{person}{Willem
  Zuidema}.} \bibinfo{year}{2020}\natexlab{}.
\newblock \showarticletitle{Quantifying attention flow in transformers}.
\newblock \bibinfo{journal}{\emph{arXiv preprint arXiv:2005.00928}}
  (\bibinfo{year}{2020}).
\newblock


\bibitem[Agarwal et~al\mbox{.}(2022)]%
        {agarwal2022semantics}
\bibfield{author}{\bibinfo{person}{Aishwarya Agarwal}, \bibinfo{person}{Biplab
  Banerjee}, \bibinfo{person}{Fabio Cuzzolin}, {and} \bibinfo{person}{Subhasis
  Chaudhuri}.} \bibinfo{year}{2022}\natexlab{}.
\newblock \showarticletitle{Semantics-Driven Generative Replay for Few-Shot
  Class Incremental Learning}. In \bibinfo{booktitle}{\emph{Proceedings of the
  30th ACM International Conference on Multimedia}}.
  \bibinfo{pages}{5246--5254}.
\newblock


\bibitem[Ahmed et~al\mbox{.}(2024)]%
        {ahmed2024orco}
\bibfield{author}{\bibinfo{person}{Noor Ahmed}, \bibinfo{person}{Anna Kukleva},
  {and} \bibinfo{person}{Bernt Schiele}.} \bibinfo{year}{2024}\natexlab{}.
\newblock \showarticletitle{OrCo: Towards Better Generalization via
  Orthogonality and Contrast for Few-Shot Class-Incremental Learning}. In
  \bibinfo{booktitle}{\emph{Proceedings of the IEEE/CVF Conference on Computer
  Vision and Pattern Recognition}}. \bibinfo{pages}{28762--28771}.
\newblock


\bibitem[Caron et~al\mbox{.}(2020)]%
        {caron2020unsupervised}
\bibfield{author}{\bibinfo{person}{Mathilde Caron}, \bibinfo{person}{Ishan
  Misra}, \bibinfo{person}{Julien Mairal}, \bibinfo{person}{Priya Goyal},
  \bibinfo{person}{Piotr Bojanowski}, {and} \bibinfo{person}{Armand Joulin}.}
  \bibinfo{year}{2020}\natexlab{}.
\newblock \showarticletitle{Unsupervised learning of visual features by
  contrasting cluster assignments}.
\newblock \bibinfo{journal}{\emph{Advances in neural information processing
  systems}}  \bibinfo{volume}{33} (\bibinfo{year}{2020}),
  \bibinfo{pages}{9912--9924}.
\newblock


\bibitem[Chi et~al\mbox{.}(2022)]%
        {chi2022metafscil}
\bibfield{author}{\bibinfo{person}{Zhixiang Chi}, \bibinfo{person}{Li Gu},
  \bibinfo{person}{Huan Liu}, \bibinfo{person}{Yang Wang},
  \bibinfo{person}{Yuanhao Yu}, {and} \bibinfo{person}{Jin Tang}.}
  \bibinfo{year}{2022}\natexlab{}.
\newblock \showarticletitle{Metafscil: A meta-learning approach for few-shot
  class incremental learning}. In \bibinfo{booktitle}{\emph{Proceedings of the
  IEEE/CVF conference on computer vision and pattern recognition}}.
  \bibinfo{pages}{14166--14175}.
\newblock


\bibitem[D'Alessandro et~al\mbox{.}(2023)]%
        {d2023multimodal}
\bibfield{author}{\bibinfo{person}{Marco D'Alessandro},
  \bibinfo{person}{Alberto Alonso}, \bibinfo{person}{Enrique Calabr{\'e}s},
  {and} \bibinfo{person}{Mikel Galar}.} \bibinfo{year}{2023}\natexlab{}.
\newblock \showarticletitle{Multimodal parameter-efficient few-shot class
  incremental learning}. In \bibinfo{booktitle}{\emph{Proceedings of the
  IEEE/CVF International Conference on Computer Vision}}.
  \bibinfo{pages}{3393--3403}.
\newblock


\bibitem[Deng et~al\mbox{.}(2009)]%
        {deng2009imagenet}
\bibfield{author}{\bibinfo{person}{Jia Deng}, \bibinfo{person}{Wei Dong},
  \bibinfo{person}{Richard Socher}, \bibinfo{person}{Li-Jia Li},
  \bibinfo{person}{Kai Li}, {and} \bibinfo{person}{Li Fei-Fei}.}
  \bibinfo{year}{2009}\natexlab{}.
\newblock \showarticletitle{Imagenet: A large-scale hierarchical image
  database}. In \bibinfo{booktitle}{\emph{2009 IEEE conference on computer
  vision and pattern recognition}}. Ieee, \bibinfo{pages}{248--255}.
\newblock


\bibitem[Derakhshani et~al\mbox{.}(2022)]%
        {derakhshani2022lifelonger}
\bibfield{author}{\bibinfo{person}{Mohammad~Mahdi Derakhshani},
  \bibinfo{person}{Ivona Najdenkoska}, \bibinfo{person}{Tom van Sonsbeek},
  \bibinfo{person}{Xiantong Zhen}, \bibinfo{person}{Dwarikanath Mahapatra},
  \bibinfo{person}{Marcel Worring}, {and} \bibinfo{person}{Cees~GM Snoek}.}
  \bibinfo{year}{2022}\natexlab{}.
\newblock \showarticletitle{Lifelonger: A benchmark for continual disease
  classification}. In \bibinfo{booktitle}{\emph{International conference on
  medical image computing and computer-assisted intervention}}. Springer,
  \bibinfo{pages}{314--324}.
\newblock


\bibitem[Doan et~al\mbox{.}(2024)]%
        {doan2024streamlined}
\bibfield{author}{\bibinfo{person}{Thang Doan}, \bibinfo{person}{Sima Behpour},
  \bibinfo{person}{Xin Li}, \bibinfo{person}{Wenbin He}, \bibinfo{person}{Liang
  Gou}, {and} \bibinfo{person}{Liu Ren}.} \bibinfo{year}{2024}\natexlab{}.
\newblock \showarticletitle{A streamlined Approach to Multimodal Few-Shot Class
  Incremental Learning for Fine-Grained Datasets}.
\newblock \bibinfo{journal}{\emph{arXiv preprint arXiv:2403.06295}}
  (\bibinfo{year}{2024}).
\newblock


\bibitem[Dosovitskiy et~al\mbox{.}(2020)]%
        {dosovitskiy2020image}
\bibfield{author}{\bibinfo{person}{Alexey Dosovitskiy}, \bibinfo{person}{Lucas
  Beyer}, \bibinfo{person}{Alexander Kolesnikov}, \bibinfo{person}{Dirk
  Weissenborn}, \bibinfo{person}{Xiaohua Zhai}, \bibinfo{person}{Thomas
  Unterthiner}, \bibinfo{person}{Mostafa Dehghani}, \bibinfo{person}{Matthias
  Minderer}, \bibinfo{person}{Georg Heigold}, \bibinfo{person}{Sylvain Gelly},
  {et~al\mbox{.}}} \bibinfo{year}{2020}\natexlab{}.
\newblock \showarticletitle{An image is worth 16x16 words: Transformers for
  image recognition at scale}.
\newblock \bibinfo{journal}{\emph{arXiv preprint arXiv:2010.11929}}
  (\bibinfo{year}{2020}).
\newblock


\bibitem[Ghiasi et~al\mbox{.}(2022)]%
        {ghiasi2022scaling}
\bibfield{author}{\bibinfo{person}{Golnaz Ghiasi}, \bibinfo{person}{Xiuye Gu},
  \bibinfo{person}{Yin Cui}, {and} \bibinfo{person}{Tsung-Yi Lin}.}
  \bibinfo{year}{2022}\natexlab{}.
\newblock \showarticletitle{Scaling open-vocabulary image segmentation with
  image-level labels}. In \bibinfo{booktitle}{\emph{European conference on
  computer vision}}. Springer, \bibinfo{pages}{540--557}.
\newblock


\bibitem[Goswami et~al\mbox{.}(2024)]%
        {goswami2024calibrating}
\bibfield{author}{\bibinfo{person}{Dipam Goswami},
  \bibinfo{person}{Bart{\l}omiej Twardowski}, {and} \bibinfo{person}{Joost Van
  De~Weijer}.} \bibinfo{year}{2024}\natexlab{}.
\newblock \showarticletitle{Calibrating Higher-Order Statistics for Few-Shot
  Class-Incremental Learning with Pre-trained Vision Transformers}. In
  \bibinfo{booktitle}{\emph{Proceedings of the IEEE/CVF Conference on Computer
  Vision and Pattern Recognition}}. \bibinfo{pages}{4075--4084}.
\newblock


\bibitem[Gu et~al\mbox{.}(2023)]%
        {gu2023few}
\bibfield{author}{\bibinfo{person}{Ziqi Gu}, \bibinfo{person}{Chunyan Xu},
  \bibinfo{person}{Jian Yang}, {and} \bibinfo{person}{Zhen Cui}.}
  \bibinfo{year}{2023}\natexlab{}.
\newblock \showarticletitle{Few-shot continual infomax learning}. In
  \bibinfo{booktitle}{\emph{Proceedings of the IEEE/CVF International
  Conference on Computer Vision}}. \bibinfo{pages}{19224--19233}.
\newblock


\bibitem[Han et~al\mbox{.}(2022)]%
        {han2022survey}
\bibfield{author}{\bibinfo{person}{Kai Han}, \bibinfo{person}{Yunhe Wang},
  \bibinfo{person}{Hanting Chen}, \bibinfo{person}{Xinghao Chen},
  \bibinfo{person}{Jianyuan Guo}, \bibinfo{person}{Zhenhua Liu},
  \bibinfo{person}{Yehui Tang}, \bibinfo{person}{An Xiao},
  \bibinfo{person}{Chunjing Xu}, \bibinfo{person}{Yixing Xu}, {et~al\mbox{.}}}
  \bibinfo{year}{2022}\natexlab{}.
\newblock \showarticletitle{A survey on vision transformer}.
\newblock \bibinfo{journal}{\emph{IEEE transactions on pattern analysis and
  machine intelligence}} \bibinfo{volume}{45}, \bibinfo{number}{1}
  (\bibinfo{year}{2022}), \bibinfo{pages}{87--110}.
\newblock


\bibitem[He et~al\mbox{.}(2016)]%
        {he2016deep}
\bibfield{author}{\bibinfo{person}{Kaiming He}, \bibinfo{person}{Xiangyu
  Zhang}, \bibinfo{person}{Shaoqing Ren}, {and} \bibinfo{person}{Jian Sun}.}
  \bibinfo{year}{2016}\natexlab{}.
\newblock \showarticletitle{Deep residual learning for image recognition}. In
  \bibinfo{booktitle}{\emph{Proceedings of the IEEE conference on computer
  vision and pattern recognition}}. \bibinfo{pages}{770--778}.
\newblock


\bibitem[Hendrycks et~al\mbox{.}(2021)]%
        {hendrycks2021many}
\bibfield{author}{\bibinfo{person}{Dan Hendrycks}, \bibinfo{person}{Steven
  Basart}, \bibinfo{person}{Norman Mu}, \bibinfo{person}{Saurav Kadavath},
  \bibinfo{person}{Frank Wang}, \bibinfo{person}{Evan Dorundo},
  \bibinfo{person}{Rahul Desai}, \bibinfo{person}{Tyler Zhu},
  \bibinfo{person}{Samyak Parajuli}, \bibinfo{person}{Mike Guo},
  {et~al\mbox{.}}} \bibinfo{year}{2021}\natexlab{}.
\newblock \showarticletitle{The many faces of robustness: A critical analysis
  of out-of-distribution generalization}. In
  \bibinfo{booktitle}{\emph{Proceedings of the IEEE/CVF international
  conference on computer vision}}. \bibinfo{pages}{8340--8349}.
\newblock


\bibitem[Hersche et~al\mbox{.}(2022)]%
        {hersche2022constrained}
\bibfield{author}{\bibinfo{person}{Michael Hersche}, \bibinfo{person}{Geethan
  Karunaratne}, \bibinfo{person}{Giovanni Cherubini}, \bibinfo{person}{Luca
  Benini}, \bibinfo{person}{Abu Sebastian}, {and} \bibinfo{person}{Abbas
  Rahimi}.} \bibinfo{year}{2022}\natexlab{}.
\newblock \showarticletitle{Constrained few-shot class-incremental learning}.
  In \bibinfo{booktitle}{\emph{CVPR}}. \bibinfo{pages}{9057--9067}.
\newblock


\bibitem[Huang et~al\mbox{.}(2024)]%
        {huang2024learning}
\bibfield{author}{\bibinfo{person}{Zitong Huang}, \bibinfo{person}{Ze Chen},
  \bibinfo{person}{Zhixing Chen}, \bibinfo{person}{Erjin Zhou},
  \bibinfo{person}{Xinxing Xu}, \bibinfo{person}{Rick Siow~Mong Goh},
  \bibinfo{person}{Yong Liu}, \bibinfo{person}{Wangmeng Zuo}, {and}
  \bibinfo{person}{Chunmei Feng}.} \bibinfo{year}{2024}\natexlab{}.
\newblock \showarticletitle{Learning prompt with distribution-based feature
  replay for few-shot class-incremental learning}.
\newblock \bibinfo{journal}{\emph{arXiv preprint arXiv:2401.01598}}
  (\bibinfo{year}{2024}).
\newblock


\bibitem[Jia et~al\mbox{.}(2022)]%
        {jia2022visual}
\bibfield{author}{\bibinfo{person}{Menglin Jia}, \bibinfo{person}{Luming Tang},
  \bibinfo{person}{Bor-Chun Chen}, \bibinfo{person}{Claire Cardie},
  \bibinfo{person}{Serge Belongie}, \bibinfo{person}{Bharath Hariharan}, {and}
  \bibinfo{person}{Ser-Nam Lim}.} \bibinfo{year}{2022}\natexlab{}.
\newblock \showarticletitle{Visual prompt tuning}. In
  \bibinfo{booktitle}{\emph{European conference on computer vision}}. Springer,
  \bibinfo{pages}{709--727}.
\newblock


\bibitem[Jia et~al\mbox{.}(2024)]%
        {jia2024purified}
\bibfield{author}{\bibinfo{person}{Shilong Jia}, \bibinfo{person}{Tingting Wu},
  \bibinfo{person}{Yingying Fang}, \bibinfo{person}{Tieyong Zeng},
  \bibinfo{person}{Guixu Zhang}, {and} \bibinfo{person}{Zhi Li}.}
  \bibinfo{year}{2024}\natexlab{}.
\newblock \showarticletitle{Purified Distillation: Bridging Domain Shift and
  Category Gap in Incremental Object Detection}. In
  \bibinfo{booktitle}{\emph{Proceedings of the 32nd ACM International
  Conference on Multimedia}}. \bibinfo{pages}{1197--1205}.
\newblock


\bibitem[Khan et~al\mbox{.}(2023)]%
        {khan2023introducing}
\bibfield{author}{\bibinfo{person}{Muhammad Gul Zain~Ali Khan},
  \bibinfo{person}{Muhammad~Ferjad Naeem}, \bibinfo{person}{Luc Van~Gool},
  \bibinfo{person}{Didier Stricker}, \bibinfo{person}{Federico Tombari}, {and}
  \bibinfo{person}{Muhammad~Zeshan Afzal}.} \bibinfo{year}{2023}\natexlab{}.
\newblock \showarticletitle{Introducing language guidance in prompt-based
  continual learning}. In \bibinfo{booktitle}{\emph{Proceedings of the IEEE/CVF
  International Conference on Computer Vision}}. \bibinfo{pages}{11463--11473}.
\newblock


\bibitem[Khattak et~al\mbox{.}(2023)]%
        {khattak2023maple}
\bibfield{author}{\bibinfo{person}{Muhammad~Uzair Khattak},
  \bibinfo{person}{Hanoona Rasheed}, \bibinfo{person}{Muhammad Maaz},
  \bibinfo{person}{Salman Khan}, {and} \bibinfo{person}{Fahad~Shahbaz Khan}.}
  \bibinfo{year}{2023}\natexlab{}.
\newblock \showarticletitle{Maple: Multi-modal prompt learning}. In
  \bibinfo{booktitle}{\emph{Proceedings of the IEEE/CVF conference on computer
  vision and pattern recognition}}. \bibinfo{pages}{19113--19122}.
\newblock


\bibitem[Krause et~al\mbox{.}(2013)]%
        {krause20133d}
\bibfield{author}{\bibinfo{person}{Jonathan Krause}, \bibinfo{person}{Michael
  Stark}, \bibinfo{person}{Jia Deng}, {and} \bibinfo{person}{Li Fei-Fei}.}
  \bibinfo{year}{2013}\natexlab{}.
\newblock \showarticletitle{3d object representations for fine-grained
  categorization}. In \bibinfo{booktitle}{\emph{Proceedings of the IEEE
  international conference on computer vision workshops}}.
  \bibinfo{pages}{554--561}.
\newblock


\bibitem[Krizhevsky et~al\mbox{.}(2009)]%
        {krizhevsky2009learning}
\bibfield{author}{\bibinfo{person}{Alex Krizhevsky}, \bibinfo{person}{Geoffrey
  Hinton}, {et~al\mbox{.}}} \bibinfo{year}{2009}\natexlab{}.
\newblock \showarticletitle{Learning multiple layers of features from tiny
  images}.
\newblock \bibinfo{journal}{\emph{Master's thesis, University of Tront}}
  (\bibinfo{year}{2009}).
\newblock


\bibitem[Li et~al\mbox{.}(2022)]%
        {li2022blip}
\bibfield{author}{\bibinfo{person}{Junnan Li}, \bibinfo{person}{Dongxu Li},
  \bibinfo{person}{Caiming Xiong}, {and} \bibinfo{person}{Steven Hoi}.}
  \bibinfo{year}{2022}\natexlab{}.
\newblock \showarticletitle{Blip: Bootstrapping language-image pre-training for
  unified vision-language understanding and generation}. In
  \bibinfo{booktitle}{\emph{International conference on machine learning}}.
  PMLR, \bibinfo{pages}{12888--12900}.
\newblock


\bibitem[Li et~al\mbox{.}(2024)]%
        {li2024few}
\bibfield{author}{\bibinfo{person}{Yanan Li}, \bibinfo{person}{Linpu He},
  \bibinfo{person}{Feng Lin}, {and} \bibinfo{person}{Donghui Wang}.}
  \bibinfo{year}{2024}\natexlab{}.
\newblock \showarticletitle{Few-Shot Class-Incremental Learning via Cross-Modal
  Alignment with Feature Replay}. In \bibinfo{booktitle}{\emph{Chinese
  Conference on Pattern Recognition and Computer Vision (PRCV)}}. Springer,
  \bibinfo{pages}{19--33}.
\newblock


\bibitem[Li et~al\mbox{.}(2025)]%
        {CMA-FSCIL}
\bibfield{author}{\bibinfo{person}{Yanan Li}, \bibinfo{person}{Linpu He},
  \bibinfo{person}{Feng Lin}, {and} \bibinfo{person}{Donghui Wang}.}
  \bibinfo{year}{2025}\natexlab{}.
\newblock \showarticletitle{Few-Shot Class-Incremental Learning via Cross-Modal
  Alignment with Feature Replay}. In \bibinfo{booktitle}{\emph{Pattern
  Recognition and Computer Vision}}. \bibinfo{publisher}{Springer Nature
  Singapore}.
\newblock


\bibitem[Liu et~al\mbox{.}(2025)]%
        {liu2025few}
\bibfield{author}{\bibinfo{person}{Chenxi Liu}, \bibinfo{person}{Zhenyi Wang},
  \bibinfo{person}{Tianyi Xiong}, \bibinfo{person}{Ruibo Chen},
  \bibinfo{person}{Yihan Wu}, \bibinfo{person}{Junfeng Guo}, {and}
  \bibinfo{person}{Heng Huang}.} \bibinfo{year}{2025}\natexlab{}.
\newblock \showarticletitle{Few-shot class incremental learning with
  attention-aware self-adaptive prompt}. In \bibinfo{booktitle}{\emph{European
  Conference on Computer Vision}}. Springer, \bibinfo{pages}{1--18}.
\newblock


\bibitem[Liu et~al\mbox{.}(2022)]%
        {liu2022few}
\bibfield{author}{\bibinfo{person}{Huan Liu}, \bibinfo{person}{Li Gu},
  \bibinfo{person}{Zhixiang Chi}, \bibinfo{person}{Yang Wang},
  \bibinfo{person}{Yuanhao Yu}, \bibinfo{person}{Jun Chen}, {and}
  \bibinfo{person}{Jin Tang}.} \bibinfo{year}{2022}\natexlab{}.
\newblock \showarticletitle{Few-shot class-incremental learning via
  entropy-regularized data-free replay}. In \bibinfo{booktitle}{\emph{European
  Conference on Computer Vision}}. Springer, \bibinfo{pages}{146--162}.
\newblock


\bibitem[Maji et~al\mbox{.}(2013)]%
        {maji2013fine}
\bibfield{author}{\bibinfo{person}{Subhransu Maji}, \bibinfo{person}{Esa
  Rahtu}, \bibinfo{person}{Juho Kannala}, \bibinfo{person}{Matthew Blaschko},
  {and} \bibinfo{person}{Andrea Vedaldi}.} \bibinfo{year}{2013}\natexlab{}.
\newblock \showarticletitle{Fine-grained visual classification of aircraft}.
\newblock \bibinfo{journal}{\emph{arXiv preprint arXiv:1306.5151}}
  (\bibinfo{year}{2013}).
\newblock


\bibitem[Mazumder et~al\mbox{.}(2021)]%
        {mazumder2021few}
\bibfield{author}{\bibinfo{person}{Pratik Mazumder}, \bibinfo{person}{Pravendra
  Singh}, {and} \bibinfo{person}{Piyush Rai}.} \bibinfo{year}{2021}\natexlab{}.
\newblock \showarticletitle{Few-shot lifelong learning}. In
  \bibinfo{booktitle}{\emph{Proceedings of the AAAI Conference on Artificial
  Intelligence}}, Vol.~\bibinfo{volume}{35}. \bibinfo{pages}{2337--2345}.
\newblock


\bibitem[Park et~al\mbox{.}(2024)]%
        {park2024pre}
\bibfield{author}{\bibinfo{person}{Keon-Hee Park}, \bibinfo{person}{Kyungwoo
  Song}, {and} \bibinfo{person}{Gyeong-Moon Park}.}
  \bibinfo{year}{2024}\natexlab{}.
\newblock \showarticletitle{Pre-trained Vision and Language Transformers Are
  Few-Shot Incremental Learners}. In \bibinfo{booktitle}{\emph{Proceedings of
  the IEEE/CVF Conference on Computer Vision and Pattern Recognition}}.
  \bibinfo{pages}{23881--23890}.
\newblock


\bibitem[Paszke et~al\mbox{.}(2019)]%
        {paszke2019pytorch}
\bibfield{author}{\bibinfo{person}{Adam Paszke}, \bibinfo{person}{Sam Gross},
  \bibinfo{person}{Francisco Massa}, \bibinfo{person}{Adam Lerer},
  \bibinfo{person}{James Bradbury}, \bibinfo{person}{Gregory Chanan},
  \bibinfo{person}{Trevor Killeen}, \bibinfo{person}{Zeming Lin},
  \bibinfo{person}{Natalia Gimelshein}, \bibinfo{person}{Luca Antiga},
  {et~al\mbox{.}}} \bibinfo{year}{2019}\natexlab{}.
\newblock \showarticletitle{Pytorch: An imperative style, high-performance deep
  learning library}.
\newblock \bibinfo{journal}{\emph{Advances in neural information processing
  systems}}  \bibinfo{volume}{32} (\bibinfo{year}{2019}).
\newblock


\bibitem[Peng et~al\mbox{.}(2022)]%
        {peng2022few}
\bibfield{author}{\bibinfo{person}{Can Peng}, \bibinfo{person}{Kun Zhao},
  \bibinfo{person}{Tianren Wang}, \bibinfo{person}{Meng Li}, {and}
  \bibinfo{person}{Brian~C Lovell}.} \bibinfo{year}{2022}\natexlab{}.
\newblock \showarticletitle{Few-shot class-incremental learning from an
  open-set perspective}. In \bibinfo{booktitle}{\emph{European Conference on
  Computer Vision}}. Springer, \bibinfo{pages}{382--397}.
\newblock


\bibitem[Qiu et~al\mbox{.}(2023)]%
        {qiu2023semantic}
\bibfield{author}{\bibinfo{person}{Wenhao Qiu}, \bibinfo{person}{Sichao Fu},
  \bibinfo{person}{Jingyi Zhang}, \bibinfo{person}{Chengxiang Lei}, {and}
  \bibinfo{person}{Qinmu Peng}.} \bibinfo{year}{2023}\natexlab{}.
\newblock \showarticletitle{Semantic-visual guided transformer for few-shot
  class-incremental learning}. In \bibinfo{booktitle}{\emph{2023 IEEE
  International Conference on Multimedia and Expo (ICME)}}. IEEE,
  \bibinfo{pages}{2885--2890}.
\newblock


\bibitem[Radford et~al\mbox{.}(2021)]%
        {radford2021learning}
\bibfield{author}{\bibinfo{person}{Alec Radford}, \bibinfo{person}{Jong~Wook
  Kim}, \bibinfo{person}{Chris Hallacy}, \bibinfo{person}{Aditya Ramesh},
  \bibinfo{person}{Gabriel Goh}, \bibinfo{person}{Sandhini Agarwal},
  \bibinfo{person}{Girish Sastry}, \bibinfo{person}{Amanda Askell},
  \bibinfo{person}{Pamela Mishkin}, \bibinfo{person}{Jack Clark},
  {et~al\mbox{.}}} \bibinfo{year}{2021}\natexlab{}.
\newblock \showarticletitle{Learning transferable visual models from natural
  language supervision}. In \bibinfo{booktitle}{\emph{International conference
  on machine learning}}. PmLR, \bibinfo{pages}{8748--8763}.
\newblock


\bibitem[Raghu et~al\mbox{.}(2021)]%
        {raghu2021vision}
\bibfield{author}{\bibinfo{person}{Maithra Raghu}, \bibinfo{person}{Thomas
  Unterthiner}, \bibinfo{person}{Simon Kornblith}, \bibinfo{person}{Chiyuan
  Zhang}, {and} \bibinfo{person}{Alexey Dosovitskiy}.}
  \bibinfo{year}{2021}\natexlab{}.
\newblock \showarticletitle{Do vision transformers see like convolutional
  neural networks?}
\newblock \bibinfo{journal}{\emph{Advances in neural information processing
  systems}}  \bibinfo{volume}{34} (\bibinfo{year}{2021}),
  \bibinfo{pages}{12116--12128}.
\newblock


\bibitem[Ravi and Larochelle(2017)]%
        {ravi2017optimization}
\bibfield{author}{\bibinfo{person}{Sachin Ravi} {and} \bibinfo{person}{Hugo
  Larochelle}.} \bibinfo{year}{2017}\natexlab{}.
\newblock \showarticletitle{Optimization as a model for few-shot learning}. In
  \bibinfo{booktitle}{\emph{International conference on learning
  representations}}.
\newblock


\bibitem[Schuhmann et~al\mbox{.}(2021)]%
        {schuhmann2021laion}
\bibfield{author}{\bibinfo{person}{Christoph Schuhmann},
  \bibinfo{person}{Richard Vencu}, \bibinfo{person}{Romain Beaumont},
  \bibinfo{person}{Robert Kaczmarczyk}, \bibinfo{person}{Clayton Mullis},
  \bibinfo{person}{Aarush Katta}, \bibinfo{person}{Theo Coombes},
  \bibinfo{person}{Jenia Jitsev}, {and} \bibinfo{person}{Aran Komatsuzaki}.}
  \bibinfo{year}{2021}\natexlab{}.
\newblock \showarticletitle{Laion-400m: Open dataset of clip-filtered 400
  million image-text pairs}.
\newblock \bibinfo{journal}{\emph{arXiv preprint arXiv:2111.02114}}
  (\bibinfo{year}{2021}).
\newblock


\bibitem[Smith et~al\mbox{.}(2023)]%
        {smith2023coda}
\bibfield{author}{\bibinfo{person}{James~Seale Smith}, \bibinfo{person}{Leonid
  Karlinsky}, \bibinfo{person}{Vyshnavi Gutta}, \bibinfo{person}{Paola
  Cascante-Bonilla}, \bibinfo{person}{Donghyun Kim}, \bibinfo{person}{Assaf
  Arbelle}, \bibinfo{person}{Rameswar Panda}, \bibinfo{person}{Rogerio Feris},
  {and} \bibinfo{person}{Zsolt Kira}.} \bibinfo{year}{2023}\natexlab{}.
\newblock \showarticletitle{Coda-prompt: Continual decomposed attention-based
  prompting for rehearsal-free continual learning}. In
  \bibinfo{booktitle}{\emph{Proceedings of the IEEE/CVF Conference on Computer
  Vision and Pattern Recognition}}. \bibinfo{pages}{11909--11919}.
\newblock


\bibitem[Tao et~al\mbox{.}(2020)]%
        {tao2020few}
\bibfield{author}{\bibinfo{person}{Xiaoyu Tao}, \bibinfo{person}{Xiaopeng
  Hong}, \bibinfo{person}{Xinyuan Chang}, \bibinfo{person}{Songlin Dong},
  \bibinfo{person}{Xing Wei}, {and} \bibinfo{person}{Yihong Gong}.}
  \bibinfo{year}{2020}\natexlab{}.
\newblock \showarticletitle{Few-shot class-incremental learning}. In
  \bibinfo{booktitle}{\emph{Proceedings of the IEEE/CVF conference on computer
  vision and pattern recognition}}. \bibinfo{pages}{12183--12192}.
\newblock


\bibitem[Tian et~al\mbox{.}(2024)]%
        {tian2024pl}
\bibfield{author}{\bibinfo{person}{Songsong Tian}, \bibinfo{person}{Lusi Li},
  \bibinfo{person}{Weijun Li}, \bibinfo{person}{Hang Ran}, \bibinfo{person}{Li
  Li}, {and} \bibinfo{person}{Xin Ning}.} \bibinfo{year}{2024}\natexlab{}.
\newblock \showarticletitle{Pl-fscil: Harnessing the power of prompts for
  few-shot class-incremental learning}.
\newblock \bibinfo{journal}{\emph{arXiv preprint arXiv:2401.14807}}
  (\bibinfo{year}{2024}).
\newblock


\bibitem[Van~der Maaten and Hinton(2008)]%
        {van2008visualizing}
\bibfield{author}{\bibinfo{person}{Laurens Van~der Maaten} {and}
  \bibinfo{person}{Geoffrey Hinton}.} \bibinfo{year}{2008}\natexlab{}.
\newblock \showarticletitle{Visualizing data using t-SNE.}
\newblock \bibinfo{journal}{\emph{Journal of machine learning research}}
  \bibinfo{volume}{9}, \bibinfo{number}{11} (\bibinfo{year}{2008}).
\newblock


\bibitem[Wah et~al\mbox{.}(2011)]%
        {wah2011caltech}
\bibfield{author}{\bibinfo{person}{Catherine Wah}, \bibinfo{person}{Steve
  Branson}, \bibinfo{person}{Peter Welinder}, \bibinfo{person}{Pietro Perona},
  {and} \bibinfo{person}{Serge Belongie}.} \bibinfo{year}{2011}\natexlab{}.
\newblock \showarticletitle{The caltech-ucsd birds-200-2011 dataset}.
\newblock \bibinfo{journal}{\emph{CNS-TR-2011-001}} (\bibinfo{year}{2011}).
\newblock


\bibitem[Wang et~al\mbox{.}(2024)]%
        {wang2024few}
\bibfield{author}{\bibinfo{person}{Qi-Wei Wang}, \bibinfo{person}{Da-Wei Zhou},
  \bibinfo{person}{Yi-Kai Zhang}, \bibinfo{person}{De-Chuan Zhan}, {and}
  \bibinfo{person}{Han-Jia Ye}.} \bibinfo{year}{2024}\natexlab{}.
\newblock \showarticletitle{Few-shot class-incremental learning via
  training-free prototype calibration}.
\newblock \bibinfo{journal}{\emph{Advances in Neural Information Processing
  Systems}}  \bibinfo{volume}{36} (\bibinfo{year}{2024}).
\newblock


\bibitem[Wang et~al\mbox{.}(2025)]%
        {wang2025approximation}
\bibfield{author}{\bibinfo{person}{Xuan Wang}, \bibinfo{person}{Zhong Ji},
  \bibinfo{person}{Xiyao Liu}, \bibinfo{person}{Yanwei Pang}, {and}
  \bibinfo{person}{Jungong Han}.} \bibinfo{year}{2025}\natexlab{}.
\newblock \showarticletitle{On the Approximation Risk of Few-Shot
  Class-Incremental Learning}. In \bibinfo{booktitle}{\emph{European Conference
  on Computer Vision}}. Springer, \bibinfo{pages}{162--178}.
\newblock


\bibitem[Wang et~al\mbox{.}(2022)]%
        {wang2022learning}
\bibfield{author}{\bibinfo{person}{Zifeng Wang}, \bibinfo{person}{Zizhao
  Zhang}, \bibinfo{person}{Chen-Yu Lee}, \bibinfo{person}{Han Zhang},
  \bibinfo{person}{Ruoxi Sun}, \bibinfo{person}{Xiaoqi Ren},
  \bibinfo{person}{Guolong Su}, \bibinfo{person}{Vincent Perot},
  \bibinfo{person}{Jennifer Dy}, {and} \bibinfo{person}{Tomas Pfister}.}
  \bibinfo{year}{2022}\natexlab{}.
\newblock \showarticletitle{Learning to prompt for continual learning}. In
  \bibinfo{booktitle}{\emph{Proceedings of the IEEE/CVF conference on computer
  vision and pattern recognition}}. \bibinfo{pages}{139--149}.
\newblock


\bibitem[Xing et~al\mbox{.}(2019)]%
        {xing2019adaptive}
\bibfield{author}{\bibinfo{person}{Chen Xing}, \bibinfo{person}{Negar
  Rostamzadeh}, \bibinfo{person}{Boris Oreshkin}, {and}
  \bibinfo{person}{Pedro~O O~Pinheiro}.} \bibinfo{year}{2019}\natexlab{}.
\newblock \showarticletitle{Adaptive cross-modal few-shot learning}.
\newblock \bibinfo{journal}{\emph{Advances in neural information processing
  systems}}  \bibinfo{volume}{32} (\bibinfo{year}{2019}).
\newblock


\bibitem[Xu et~al\mbox{.}(2024)]%
        {xu2024clip}
\bibfield{author}{\bibinfo{person}{Yuqiao Xu}, \bibinfo{person}{Shucheng
  Huang}, {and} \bibinfo{person}{Haoliang Zhou}.}
  \bibinfo{year}{2024}\natexlab{}.
\newblock \showarticletitle{CA-CLIP: category-aware adaptation of CLIP model
  for few-shot class-incremental learning}.
\newblock \bibinfo{journal}{\emph{Multimedia Systems}} \bibinfo{volume}{30},
  \bibinfo{number}{3} (\bibinfo{year}{2024}), \bibinfo{pages}{1--14}.
\newblock


\bibitem[Yang et~al\mbox{.}(2021)]%
        {yang2021learnable}
\bibfield{author}{\bibinfo{person}{Boyu Yang}, \bibinfo{person}{Mingbao Lin},
  \bibinfo{person}{Binghao Liu}, \bibinfo{person}{Mengying Fu},
  \bibinfo{person}{Chang Liu}, \bibinfo{person}{Rongrong Ji}, {and}
  \bibinfo{person}{Qixiang Ye}.} \bibinfo{year}{2021}\natexlab{}.
\newblock \showarticletitle{Learnable expansion-and-compression network for
  few-shot class-incremental learning}.
\newblock \bibinfo{journal}{\emph{arXiv preprint arXiv:2104.02281}}
  (\bibinfo{year}{2021}).
\newblock


\bibitem[Yang et~al\mbox{.}(2022)]%
        {yang2022dynamic}
\bibfield{author}{\bibinfo{person}{Boyu Yang}, \bibinfo{person}{Mingbao Lin},
  \bibinfo{person}{Yunxiao Zhang}, \bibinfo{person}{Binghao Liu},
  \bibinfo{person}{Xiaodan Liang}, \bibinfo{person}{Rongrong Ji}, {and}
  \bibinfo{person}{Qixiang Ye}.} \bibinfo{year}{2022}\natexlab{}.
\newblock \showarticletitle{Dynamic support network for few-shot class
  incremental learning}.
\newblock \bibinfo{journal}{\emph{IEEE Transactions on Pattern Analysis and
  Machine Intelligence}} \bibinfo{volume}{45}, \bibinfo{number}{3}
  (\bibinfo{year}{2022}), \bibinfo{pages}{2945--2951}.
\newblock


\bibitem[Yang et~al\mbox{.}(2023)]%
        {yang2023neural}
\bibfield{author}{\bibinfo{person}{Yibo Yang}, \bibinfo{person}{Haobo Yuan},
  \bibinfo{person}{Xiangtai Li}, \bibinfo{person}{Zhouchen Lin},
  \bibinfo{person}{Philip Torr}, {and} \bibinfo{person}{Dacheng Tao}.}
  \bibinfo{year}{2023}\natexlab{}.
\newblock \showarticletitle{Neural collapse inspired feature-classifier
  alignment for few-shot class incremental learning}.
\newblock \bibinfo{journal}{\emph{arXiv preprint arXiv:2302.03004}}
  (\bibinfo{year}{2023}).
\newblock


\bibitem[Yao et~al\mbox{.}(2023)]%
        {yao2023visual}
\bibfield{author}{\bibinfo{person}{Hantao Yao}, \bibinfo{person}{Rui Zhang},
  {and} \bibinfo{person}{Changsheng Xu}.} \bibinfo{year}{2023}\natexlab{}.
\newblock \showarticletitle{Visual-language prompt tuning with knowledge-guided
  context optimization}. In \bibinfo{booktitle}{\emph{Proceedings of the
  IEEE/CVF conference on computer vision and pattern recognition}}.
  \bibinfo{pages}{6757--6767}.
\newblock


\bibitem[Yu et~al\mbox{.}(2024)]%
        {yu2024pffaa}
\bibfield{author}{\bibinfo{person}{Zhidong Yu}, \bibinfo{person}{Zhenbo Shi},
  \bibinfo{person}{Xiaoman Liu}, {and} \bibinfo{person}{Wei Yang}.}
  \bibinfo{year}{2024}\natexlab{}.
\newblock \showarticletitle{PFFAA: Prototype-based Feature and Frequency
  Alteration Attack for Semantic Segmentation}. In
  \bibinfo{booktitle}{\emph{Proceedings of the 32nd ACM International
  Conference on Multimedia}}. \bibinfo{pages}{4562--4571}.
\newblock


\bibitem[Zhang et~al\mbox{.}(2021)]%
        {zhang2021few}
\bibfield{author}{\bibinfo{person}{Chi Zhang}, \bibinfo{person}{Nan Song},
  \bibinfo{person}{Guosheng Lin}, \bibinfo{person}{Yun Zheng},
  \bibinfo{person}{Pan Pan}, {and} \bibinfo{person}{Yinghui Xu}.}
  \bibinfo{year}{2021}\natexlab{}.
\newblock \showarticletitle{Few-shot incremental learning with continually
  evolved classifiers}. In \bibinfo{booktitle}{\emph{Proceedings of the
  IEEE/CVF conference on computer vision and pattern recognition}}.
  \bibinfo{pages}{12455--12464}.
\newblock


\bibitem[Zhang et~al\mbox{.}(2025)]%
        {zhang2025few}
\bibfield{author}{\bibinfo{person}{Jinghua Zhang}, \bibinfo{person}{Li Liu},
  \bibinfo{person}{Olli Silv{\'e}n}, \bibinfo{person}{Matti Pietik{\"a}inen},
  {and} \bibinfo{person}{Dewen Hu}.} \bibinfo{year}{2025}\natexlab{}.
\newblock \showarticletitle{Few-Shot Class-Incremental Learning for
  Classification and Object Detection: A Survey}.
\newblock \bibinfo{journal}{\emph{IEEE Transactions on Pattern Analysis and
  Machine Intelligence}} (\bibinfo{year}{2025}).
\newblock


\bibitem[Zhang et~al\mbox{.}(2023)]%
        {zhang2023prompt}
\bibfield{author}{\bibinfo{person}{Renrui Zhang}, \bibinfo{person}{Xiangfei
  Hu}, \bibinfo{person}{Bohao Li}, \bibinfo{person}{Siyuan Huang},
  \bibinfo{person}{Hanqiu Deng}, \bibinfo{person}{Yu Qiao},
  \bibinfo{person}{Peng Gao}, {and} \bibinfo{person}{Hongsheng Li}.}
  \bibinfo{year}{2023}\natexlab{}.
\newblock \showarticletitle{Prompt, generate, then cache: Cascade of foundation
  models makes strong few-shot learners}. In
  \bibinfo{booktitle}{\emph{Proceedings of the IEEE/CVF conference on computer
  vision and pattern recognition}}. \bibinfo{pages}{15211--15222}.
\newblock


\bibitem[Zhao et~al\mbox{.}(2023)]%
        {zhao2023few}
\bibfield{author}{\bibinfo{person}{Linglan Zhao}, \bibinfo{person}{Jing Lu},
  \bibinfo{person}{Yunlu Xu}, \bibinfo{person}{Zhanzhan Cheng},
  \bibinfo{person}{Dashan Guo}, \bibinfo{person}{Yi Niu}, {and}
  \bibinfo{person}{Xiangzhong Fang}.} \bibinfo{year}{2023}\natexlab{}.
\newblock \showarticletitle{Few-shot class-incremental learning via class-aware
  bilateral distillation}. In \bibinfo{booktitle}{\emph{Proceedings of the
  IEEE/CVF conference on computer vision and pattern recognition}}.
  \bibinfo{pages}{11838--11847}.
\newblock


\bibitem[Zhou et~al\mbox{.}(2024)]%
        {zhou2024revisiting}
\bibfield{author}{\bibinfo{person}{Da-Wei Zhou}, \bibinfo{person}{Zi-Wen Cai},
  \bibinfo{person}{Han-Jia Ye}, \bibinfo{person}{De-Chuan Zhan}, {and}
  \bibinfo{person}{Ziwei Liu}.} \bibinfo{year}{2024}\natexlab{}.
\newblock \showarticletitle{Revisiting class-incremental learning with
  pre-trained models: Generalizability and adaptivity are all you need}.
\newblock \bibinfo{journal}{\emph{International Journal of Computer Vision}}
  (\bibinfo{year}{2024}), \bibinfo{pages}{1--21}.
\newblock


\bibitem[Zhou et~al\mbox{.}(2022a)]%
        {zhou2022forward}
\bibfield{author}{\bibinfo{person}{Da-Wei Zhou}, \bibinfo{person}{Fu-Yun Wang},
  \bibinfo{person}{Han-Jia Ye}, \bibinfo{person}{Liang Ma},
  \bibinfo{person}{Shiliang Pu}, {and} \bibinfo{person}{De-Chuan Zhan}.}
  \bibinfo{year}{2022}\natexlab{a}.
\newblock \showarticletitle{Forward compatible few-shot class-incremental
  learning}. In \bibinfo{booktitle}{\emph{Proceedings of the IEEE/CVF
  conference on computer vision and pattern recognition}}.
  \bibinfo{pages}{9046--9056}.
\newblock


\bibitem[Zhou et~al\mbox{.}(2022d)]%
        {zhou2022few}
\bibfield{author}{\bibinfo{person}{Da-Wei Zhou}, \bibinfo{person}{Han-Jia Ye},
  \bibinfo{person}{Liang Ma}, \bibinfo{person}{Di Xie},
  \bibinfo{person}{Shiliang Pu}, {and} \bibinfo{person}{De-Chuan Zhan}.}
  \bibinfo{year}{2022}\natexlab{d}.
\newblock \showarticletitle{Few-shot class-incremental learning by sampling
  multi-phase tasks}.
\newblock \bibinfo{journal}{\emph{IEEE Transactions on Pattern Analysis and
  Machine Intelligence}} \bibinfo{volume}{45}, \bibinfo{number}{11}
  (\bibinfo{year}{2022}), \bibinfo{pages}{12816--12831}.
\newblock


\bibitem[Zhou et~al\mbox{.}(2022b)]%
        {zhou2022conditional}
\bibfield{author}{\bibinfo{person}{Kaiyang Zhou}, \bibinfo{person}{Jingkang
  Yang}, \bibinfo{person}{Chen~Change Loy}, {and} \bibinfo{person}{Ziwei Liu}.}
  \bibinfo{year}{2022}\natexlab{b}.
\newblock \showarticletitle{Conditional prompt learning for vision-language
  models}. In \bibinfo{booktitle}{\emph{Proceedings of the IEEE/CVF conference
  on computer vision and pattern recognition}}. \bibinfo{pages}{16816--16825}.
\newblock


\bibitem[Zhou et~al\mbox{.}(2022c)]%
        {zhou2022learning}
\bibfield{author}{\bibinfo{person}{Kaiyang Zhou}, \bibinfo{person}{Jingkang
  Yang}, \bibinfo{person}{Chen~Change Loy}, {and} \bibinfo{person}{Ziwei Liu}.}
  \bibinfo{year}{2022}\natexlab{c}.
\newblock \showarticletitle{Learning to prompt for vision-language models}.
\newblock \bibinfo{journal}{\emph{International Journal of Computer Vision}}
  \bibinfo{volume}{130}, \bibinfo{number}{9} (\bibinfo{year}{2022}),
  \bibinfo{pages}{2337--2348}.
\newblock


\end{thebibliography}







\end{document}